\documentclass[runningheads]{llncs}

\usepackage{eccv}

\usepackage{eccvabbrv}

\usepackage{graphicx}
\usepackage{booktabs}

\usepackage{colortbl}
\usepackage{multirow}
\usepackage{multicol}
\usepackage{wrapfig}

\usepackage[accsupp]{axessibility}  

\usepackage{hyperref}

\usepackage{orcidlink}

\begin{document}

\title{Agent Attention: On the Integration of Softmax and Linear Attention} 

\titlerunning{Agent Attention}

\author{
Dongchen Han\inst{1}$\thanks{Equal contribution.}$\orcidlink{0009-0009-3431-6189} \and
Tianzhu Ye\inst{1}$^\star$\orcidlink{0009-0003-9772-4768} \and
Yizeng Han\inst{1}\orcidlink{0000-0001-5706-8784} \and
Zhuofan Xia\inst{1}\orcidlink{0009-0001-7965-364X} \and
Siyuan Pan\inst{2}\orcidlink{0000-0003-4426-8706} \and
Pengfei Wan\inst{2}\orcidlink{0000-0001-7225-565X} \and
Shiji Song\inst{1}\orcidlink{0000-0001-7361-9283} \and
Gao Huang\inst{1}$\thanks{Corresponding Author.}$\orcidlink{0000-0002-7251-0988}
}

\authorrunning{Han et al.}

\institute{Department of Automation, Tsinghua University \and Kuaishou Technology}

\maketitle

\begin{abstract}

The attention module is the key component in Transformers. While the global attention mechanism offers high expressiveness, its excessive computational cost restricts its applicability in various scenarios. In this paper, we propose a novel attention paradigm, \textbf{Agent Attention}, to strike a favorable balance between computational efficiency and representation power. Specifically, the Agent Attention, denoted as a quadruple $(Q, A, K, V)$, introduces an additional set of agent tokens $A$ into the conventional attention module.  The agent tokens first act as the agent for the query tokens $Q$ to aggregate information from $K$ and $V$, and then broadcast the information back to $Q$. Given the number of agent tokens can be designed to be much smaller than the number of query tokens, agent attention is significantly more efficient than the widely adopted Softmax attention, while preserving global context modelling capability. Interestingly, we show that the proposed agent attention is equivalent to a generalized form of linear attention. Therefore, agent attention seamlessly integrates the powerful Softmax attention and the highly efficient linear attention. Extensive experiments demonstrate the effectiveness of agent attention with various vision Transformers and across diverse vision tasks, including image classification, object detection, semantic segmentation and image generation. Notably, agent attention has shown remarkable performance in high-resolution scenarios, owning to its linear attention nature. For instance, when applied to Stable Diffusion, our agent attention accelerates generation and substantially enhances image generation quality without any additional training. 
Code is available at \href{https://github.com/LeapLabTHU/Agent-Attention}{https://github.com/LeapLabTHU/Agent-Attention}.


\end{abstract}
    
\section{Introduction}
\label{sec:intro}

Originating from natural language processing, Transformer models have rapidly gained prominence in the field of computer vision in recent years, achieving significant success in image classification~\cite{vit, deit,han2022learning,han2023dynamic}, object detection~\cite{pvt, detr}, semantic segmentation~\cite{maskformer,segformer}, and multimodal tasks~\cite{clip, knowledge_clip, gsva}. 

Nevertheless, incorporating Transformers and self-attention into the visual domain presents formidable challenges. Modern Transformer models commonly employ Softmax attention~\cite{attention}, which computes the similarity between each query-key pair, resulting in quadratic computation complexity with respect to the number of tokens. As a result, directly applying Softmax attention with global receptive fields to the visual tasks can lead to unmanageable computational demands. To tackle this issue, existing works~\cite{pvt, swin, dat, nat, biformer,han2024latency,han2021dynamic} attempt to reduce computation complexity by designing efficient attention patterns. As two representatives, Swin Transformer~\cite{swin} reduces the receptive field and confines self-attention calculations to local windows. PVT~\cite{pvt} employs a sparse attention pattern to alleviate the computational burden by reducing the number of keys and values. Despite their effectiveness, these methods inevitably compromise the capability to model long-range relationships, and are still inferior to global self-attention mechanism.



\begin{figure}[t]
    \centering
    \includegraphics[width=0.82\linewidth]{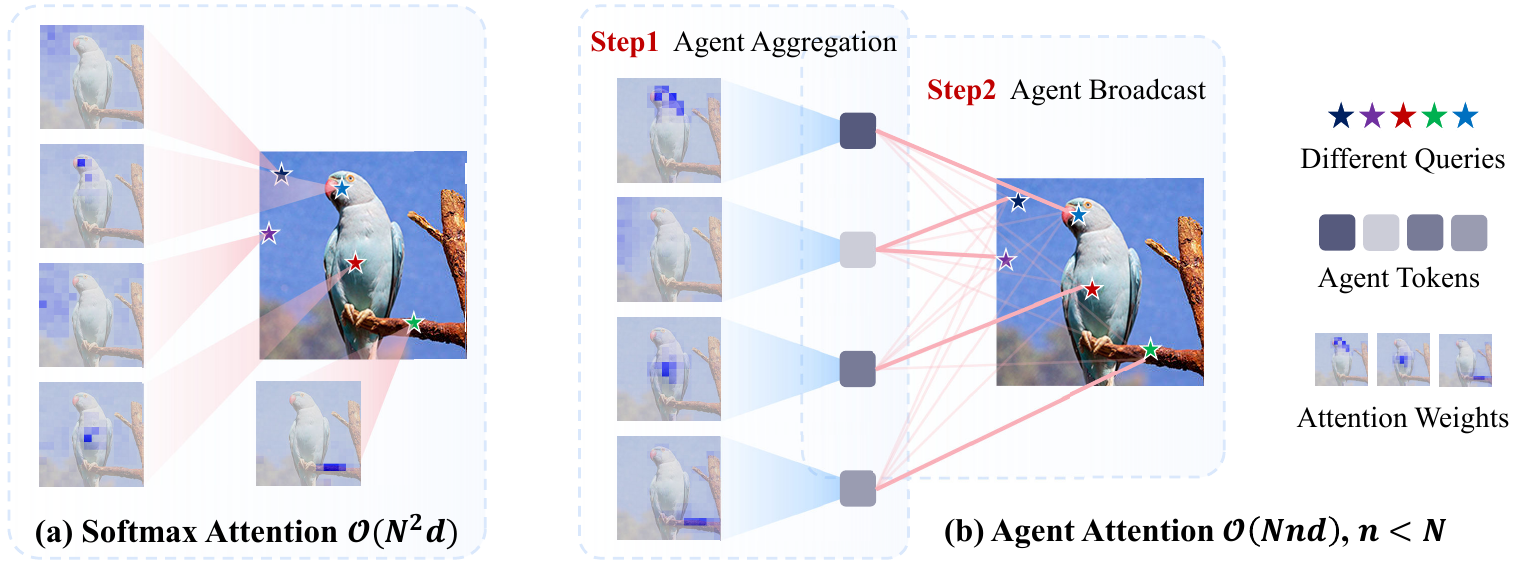}
    \caption{\textbf{An illustration of the motivation of our agent attention.} (a) In Softmax attention, each query aggregates information from all features, incurring quadratic complexity. (b) Leveraging the redundancy between attention weights, agent attention uses a small number of agent tokens to act as the ``agent'' for queries, capturing diverse semantic information from all features, and then presenting it to each query. The attention weights are derived from DeiT-T and Agent-DeiT-T.}
    \label{fig:motivation}
\end{figure}

\begin{figure}[t]
    \centering
    \includegraphics[width=0.97\linewidth]{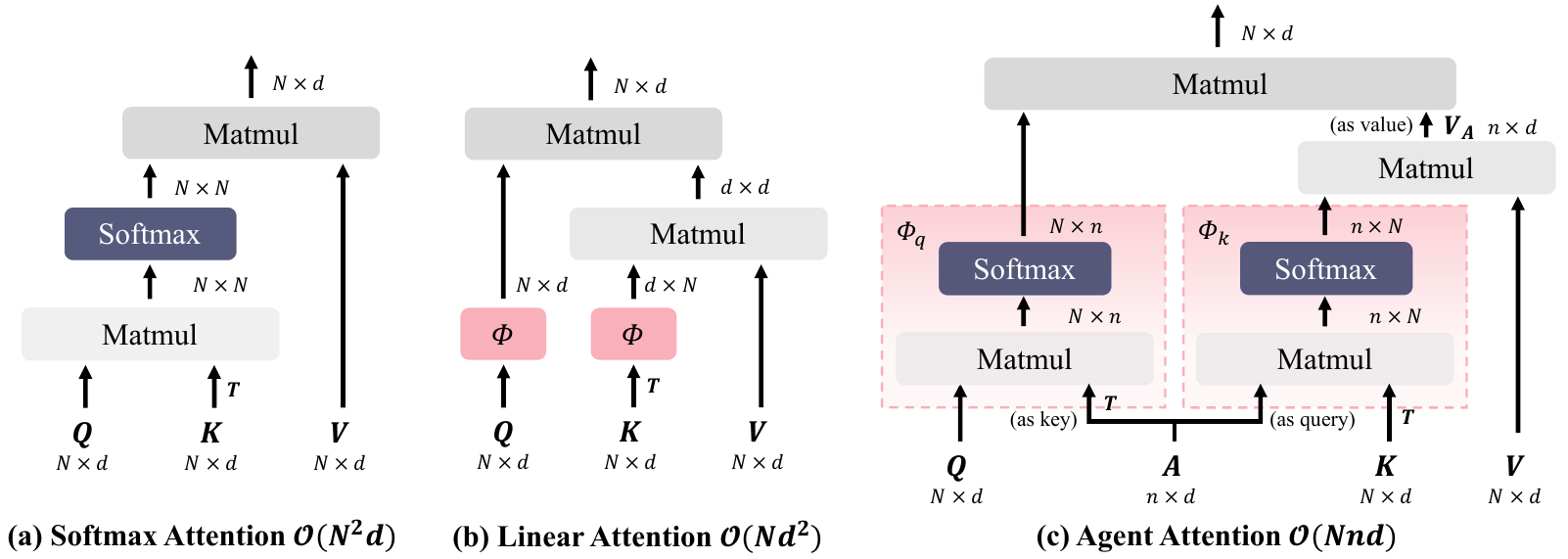}
    \caption{\textbf{Difference between Softmax attention, Linear attention and Agent attention.} (a) Softmax attention computes the similarities between all query-key pairs, resulting in quadratic complexity. (b) Linear attention applies mapping function $\phi(\cdot)$ to $Q$ and $K$ respectively to change the computation order, reducing complexity but suffering from insufficient expressive capability. (c) Our Agent attention employs a small group of agent tokens to aggregate and broadcast global information, leading to an elegant integration of Softmax and linear attention and naturally enjoying the advantages of both high expressiveness and low computation complexity.}
    \label{fig:3type_attn}
\end{figure}

In this paper, in contrast to restricting receptive field or introducing sparsity, we propose a novel quadruplet attention paradigm $(Q, A, K, V)$, dubbed \textbf{Agent Attention}, which exploits redundancy between attention weights to achieve both high model expressiveness and low computation complexity. As shown in \cref{fig:motivation}, in Softmax attention, each query aggregates information from all features, incurring quadratic complexity. In fact, many queries, such as those denoting sky in \cref{fig:motivation}a, require similar information. Therefore, our motivation is to eliminate the direct contact between each query and key, and instead use a small number of agent tokens $A$ to act as the ``agent'' for queries, capturing diverse semantic information from all features, and then presenting it to each query. 
As illustrated in \cref{fig:motivation}b and \cref{fig:3type_attn}c, the resulting agent attention is composed of two conventional Softmax attention operations. The first Softmax attention treats agent tokens $A$ as \emph{queries} to aggregate agent features $V_A$ from all values $V$, and the second utilizes agent tokens $A$ as \emph{keys}, broadcasting the global information from agent features $V_A$ to each query and forming the final output. Intuitively, the newly introduced tokens $A$ serve as ``agent'' for the query tokens $Q$, as they directly collect information from $K$ and $V$, and then deliver the result to $Q$. 

Due to the intrinsic redundancy in global self-attention, the number of agent tokens can be designed to be much smaller than the number of query tokens. This property endows agent attention with high efficiency, reducing the quadratic complexity of Softmax attention to linear complexity while preserving global context modelling capability. Interestingly, as illustrated in  \cref{fig:3type_attn}, the proposed agent attention practically forms an elegant integration of Softmax and linear attention, which explains how it achieves both high efficiency and high expressiveness from a novel perspective.

We empirically verify the effectiveness of our model across diverse vision tasks, including image classification, object detection, semantic segmentation and image generation. Our method yields substantial improvements in various tasks, particularly in high-resolution scenarios. Noteworthy, our agent attention can be directly plugged into pre-trained large diffusion models, and without any additional training, it not only accelerates the generation process, but also notably improves the generation quality.

\section{Related Works}
\label{sec:related_works}

\noindent \textbf{Vision Transformer.}
Since the inception of Vision Transformer~\cite{vit}, self-attention has made notable strides in the realm of computer vision. However, the quadratic complexity of the prevalent Softmax attention~\cite{attention} poses a challenge in applying self-attention to visual tasks. Previous works proposed various remedies for this computational challenge. PVT~\cite{pvt} introduces sparse global attention, curbing computation cost by reducing the resolution of $K$ and $V$. Swin Transformer~\cite{swin} restricts self-attention computations to local windows and employs shifted windows to model the entire image. NAT~\cite{nat} emulates convolutional operations and calculates attention within the neighborhood of each feature. DAT~\cite{dat} designs a deformable attention module to achieve a data-dependent attention pattern. BiFormer~\cite{biformer} uses bi-level routing attention to dynamically determine areas of interest for each query. GRL~\cite{grl} employs a mixture of anchored stripe attention, window attention, and channel attention to achieve efficient image restoration.
However, these approaches inherently limit the global receptive field of self-attention or are vulnerable to specifically designed attention patterns, hindering their plug-and-play adaptability for general purposes.

\noindent \textbf{Linear Attention.}
In contrast to the idea of restricting receptive fields, linear attention directly addresses the computational challenge by reducing computation complexity. The pioneer work~\cite{linear_attn} discards the Softmax function and replaces it with a mapping function $\phi$ applied to $Q$ and $K$, thereby reducing the computation complexity to $\mathcal{O}(N)$. However, such approximations led to substantial performance degradation. To tackle this problem, Efficient Attention~\cite{efficient_attn} applies the Softmax function to both $Q$ and $K$. SOFT~\cite{soft} and Nyströmformer~\cite{nystromformer} employ matrix decomposition to further approximate Softmax operation. Castling-ViT~\cite{castling_vit} uses Softmax attention as an auxiliary training tool and fully employs linear attention during inference. FLatten Transformer~\cite{flatten} proposes focused function and adopts depthwise convolution to preserve feature diversity.
While these methods are effective, they continue to struggle with the limited expressive power of linear attention. In the paper, rather than enhancing Softmax or linear attention, we propose agent attention which integrates these two attention types, achieving superior performance in various tasks.

\section{Preliminaries}
\label{sec:preliminaries}

In this section, we first review the general form of self-attention in vision Transformers and briefly analyze the pros and cons of Softmax and linear attention.

\subsection{General Form of Self-Attention}

With an input of $N$ tokens represented as $ x\in\mathbb{R}^{N \times C} $, self-attention can be formulated as follows in each head:
\begin{equation} \label{eq:general_attn}
    \begin{split}
        Q=xW_Q, K=xW_K, V=xW_V, \ \ 
        O_i=\sum_{j=1}^{N}\ \frac{{\rm Sim}{\left(Q_i,K_j\right)}}{\sum_{j=1}^{N}\ {\rm Sim}{\left(Q_i,K_j\right)}}V_j,
    \end{split}
\end{equation}
where $ W_{Q/K/V}\!\in\!\mathbb{R}^{C \times d} $ are projection matrices, $C$ and $d$ are the channel dimension of module and each head, and $ {\rm Sim}{\left(\cdot,\cdot \right)} $ denotes the similarity function.

\subsection{Softmax Attention and Linear Attention}

When using $ {\rm Sim}\left(Q,K\right)\!=\!{\rm exp}({QK^T}/{\sqrt d}) $ in \cref{eq:general_attn}, it becomes Softmax attention~\cite{attention}, which has been highly successful in modern vision Transformer designs. However, Softmax attention compels to compute the similarity between all query-key pairs, resulting in $\mathcal{O}(N^2)$ complexity. 
Consequently, using Softmax attention with a global receptive field leads to overwhelming computation complexity. To tackle this problem, previous works attempted to reduce the number of tokens $N$ by designing sparse global attention~\cite{pvt, pvtv2} or window attention~\cite{swin, cswin} patterns. While effective, these strategies unavoidably compromise the self-attention's capability for long-range modeling.

Comparably, linear attention~\cite{linear_attn} efficiently addresses the computation challenge with a linear complexity of $\mathcal{O}(N)$. Specifically, carefully designed mapping functions are applied to $Q$ and $K$ respectively, \textit{i.e.}, ${\rm Sim}\left(Q,K\right)=\phi(Q)\phi(K)^T$.
This gives us the opportunity to change the computation order from $(\phi(Q)\phi(K)^T)V$ to $\phi(Q)(\phi(K)^TV)$ based on the associative property of matrix multiplication.
As illustrated in \cref{fig:3type_attn}, by doing so, the computation complexity with respect to token number is reduced to $ \mathcal{O}(N)$.
However, designing effective mapping function $\phi(\cdot)$ proves to be a nontrivial task. Simple functions~\cite{efficient_attn} such as ReLU lead to significant performance drop, whereas more intricate designs~\cite{performer} or matrix decomposition methods~\cite{soft, nystromformer} may introduce extra computation overhead. In general, current linear attention approaches are still inferior to Softmax attention, limiting their practical application.

\begin{figure}[t]
    \centering
    \includegraphics[width=\linewidth]{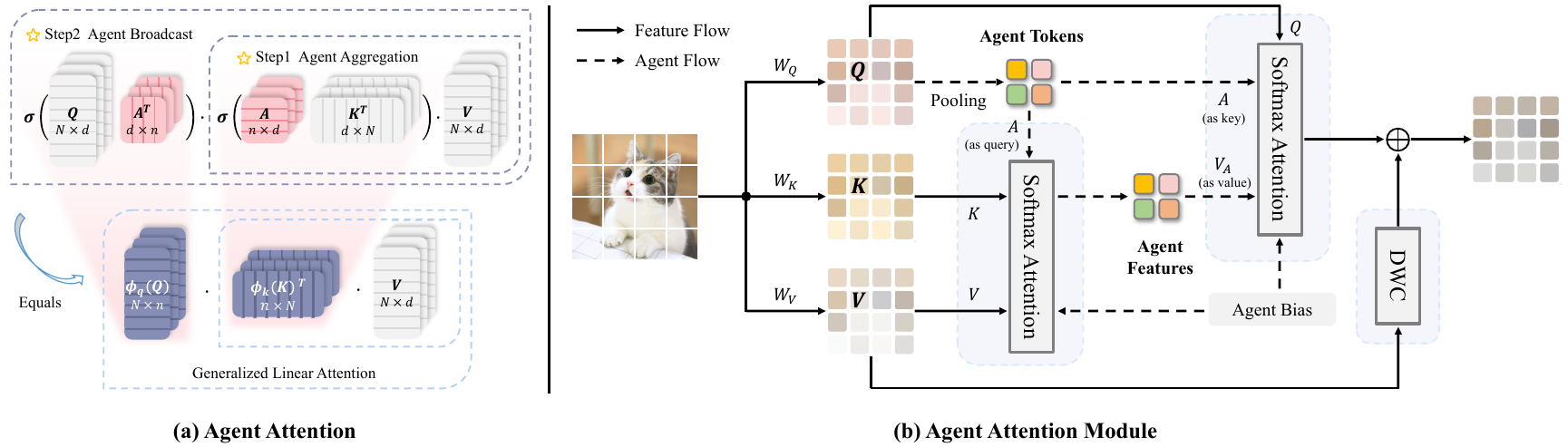}
    \caption{\textbf{
    An illustration of our agent attention and agent attention module.} (a) Agent attention uses agent tokens to aggregate global information and distribute it to individual image tokens, resulting in a practical integration of Softmax and linear attention. $\rm{\sigma}(\cdot)$ represents Softmax function. In (b), we depict the information flow of agent attention module. As a showcase, we acquire agent tokens through pooling. Subsequently, agent tokens are utilized to aggregate information from $V$, and $Q$ queries features from the agent features. In addition, agent bias and DWC are adopted to add positional information and maintain feature diversity.}
    \label{fig:agent_attn_module}
\end{figure}

\section{Agent Transformer}
\label{sec:agent_transformer}

As discussed in \cref{sec:preliminaries}, Softmax and linear attention suffer from either excessive computation complexity or insufficient model expressiveness. Previous research commonly treated these two attention paradigms as distinct approaches and attempted to either reduce the computation cost of Softmax attention or enhance the performance of linear attention. In this section, we propose a new attention paradigm named \textbf{Agent Attention}, which practically forms an elegant integration of Softmax and linear attention, enjoying benefits from both linear complexity and high expressiveness.

\subsection{Agent Attention} \label{sec:agent_attn}

To simplify, we abbreviate Softmax and linear attention as:
\begin{equation}
    \begin{split}
        O^{\rm S}\!=\!{\rm \sigma}(QK^T)V \triangleq {\rm Attn}^{\rm S}(Q,K,V), \ 
        O^{\rm \phi}\!=\!\phi(Q)\phi(K)^TV \triangleq {\rm Attn}^{\rm \phi}(Q,K,V), 
    \end{split}
\end{equation}
where $ Q, K, V\in\mathbb{R}^{N \times C} $ denote query, key and value matrices and $\rm{\sigma}(\cdot)$ represents Softmax function.
Then our agent attention can be written as:
\begin{equation} \label{eq:agnet_use_two_softmax}
    \begin{split}
        O^{\rm A}=&\ \underbrace{{\rm Attn}^{\rm S}(Q,A,\underbrace{{\rm Attn}^{\rm S}(A,K,V)}_{\rm Agent\ Aggregation})}_{\rm Agent\ Broadcast}. \\
    \end{split}
\end{equation}
It is equivalent to:
\begin{equation} \label{eq:agent_is_linear}
    \begin{split}
        O^{\rm A}=\ {\rm \sigma}(QA^T)\ {\rm \sigma}(AK^T)\ V 
        =\ {\phi}_q(Q){\phi}_k(K)^TV 
        =\ \underbrace{{\rm Attn}^{\rm \phi_{q/k}}(Q,K,V)}_{\rm Generalized\ Linear\ Attn}, \\
    \end{split}
\end{equation}
where $ A\in\mathbb{R}^{n \times C} $ is our newly defined agent tokens.


As shown in \cref{eq:agnet_use_two_softmax} and \cref{fig:agent_attn_module}a, our agent attention consists of two Softmax attention operations, namely agent aggregation and agent broadcast. Specifically, we initially treat agent tokens $A$ as \textit{queries} and perform attention calculations between $A$, $K$, and $V$ to aggregate agent features $V_A$ from all values. Subsequently, we utilize $A$ as \textit{keys} and $V_A$ as \textit{values} in the second attention calculation with the query matrix $Q$, broadcasting the global information from agent features to every query token and obtaining the final output $O$. 
In this way, we avoid the computation of pairwise similarities between $Q$ and $K$ while preserving information exchange between each query-key pair through agent tokens.


The newly defined agent tokens $A$ essentially serve as the \textit{agent} for $Q$, aggregating global information from $K$ and $V$, and subsequently broadcasting it back to $Q$. Practically, we set the number of agent tokens $n$ as a small hyper-parameter, achieving a linear complexity of $\mathcal{O}(Nnd)$ relative to the number of input features $N$ while maintaining global context modeling capability.
Interestingly, as shown in \cref{eq:agent_is_linear} and \cref{fig:agent_attn_module}a, we practically integrate the powerful Softmax attention and efficient linear attention, establishing a generalized linear attention paradigm by employing two Softmax attention operations, with the equivalent mapping function defined as $ {\phi}_{q}(Q)={\rm \sigma}(QA^T), {\phi}_{k}(K)={\left({\rm \sigma}(AK^T)\right)}^T $.

In practice, agent tokens can be acquired through different methods, such as simply setting as a set of learnable parameters or extracting from input features through pooling or convolution. It is worth noticing that more advanced techniques like deformed points~\cite{dat} or token merging~\cite{tome} can also be used to obtain agent tokens. In the default setting, we employ the simple pooling strategy to obtain agent tokens, which already works surprisingly well.

\subsection{Agent Attention Module}

Agent attention inherits the merits of both Softmax and linear attention. In practical use, we further make two improvements to maximize its potential.

\noindent \textbf{Agent Bias.}
In order to better utilize positional information, we present a carefully designed \textit{Agent Bias} for our agent attention. Specifically, inspired by RPB~\cite{rpe}, we introduce agent bias within the attention calculation, i.e.,
\begin{equation} \label{eq:agnet_bias}
    \begin{split}
        O^{\rm A}\!=\ &{\rm \sigma}(QA^T\!\!+\!B_2)\ {\rm \sigma}(AK^T\!\!+\!B_1)\ V, \\
    \end{split}
\end{equation}
where $ B_1\in\mathbb{R}^{n \times N},B_2\in\mathbb{R}^{N \times n} $ are our agent biases. 
For parameter efficiency, we construct each agent bias using three bias components rather than directly setting $B_1, B_2$ as learnable parameters (see Appendix). Agent bias augments the vanilla agent attention with spatial information, helping different agent tokens to focus on diverse regions. As shown in \cref{tab:ablation_agent_bias_dwc}, significant improvements can be observed upon the introduction of our agent bias terms.



\noindent \textbf{Diversity Restoration Module.}
Although agent attention benefits from both low computation cost and high expressiveness, as generalized linear attention, it also suffers from insufficient feature diversity~\cite{flatten}. As a remedy, we follow \cite{flatten} and adopt a depthwise convolution (DWC) module to preserve feature diversity.



\noindent \textbf{Agent Attention Module.}
Building upon these designs, we propose a novel attention module named \textit{Agent Attention Module}. As illustrated in \cref{fig:agent_attn_module}(b), our module is composed of three parts, namely pure agent attention, agent bias and the DWC module. Our module can be formulated as:
\begin{equation} \label{eq:agent_attn_module1}
    \begin{split}
        O=&\ {\rm \sigma}(QA^T\!\!+\!B_2)\ {\rm \sigma}(AK^T\!\!+\!B_1)\ V+{\rm DWC}(V),
    \end{split}
\end{equation}
where $Q,K,V\in\mathbb{R}^{N \times C}$, $A\!\in\!\mathbb{R}^{n \times C}$, $B_1\!\in\!\mathbb{R}^{n \times N}$ and $B_2\!\in\!\mathbb{R}^{N \times n}$.
In the default setting, agent tokens $A$ is obtained through pooling, i.e., $A={\rm Pooling}(Q)$.
The overall module complexity is expressed as: 
\begin{equation} \label{eq:complexity}
    \begin{split}
        \Omega =&\ \underbrace{4NC^2}_{\rm Proj}\ +\!\!\!\!\!\underbrace{NC}_{\rm Get\ Agents}\!\!\!\!\!+\ \underbrace{2nNC+2NnC}_{\rm Agent\ Attention}\ +\  \underbrace{k^2NC}_{\rm DWC},
    \end{split}
\end{equation}
where $N, n$ are the number of input features and agent tokens, and $k=3$ is the kernel size of DWC. Notably, our model exhibits linear complexity for $N$.

Combining the merits of Softmax and linear attention, our module offers the following advantages:

(1) \textbf{Efficient computation and high expressive capability}. Previous work usually viewed Softmax attention and linear attention as two different attention paradigms, aiming to address their respective limitations. As a seamless integration of these two attention forms, our agent attention naturally inherits the merits of the two, enjoying both low computation complexity and high model expression ability at the same time.

(2) \textbf{Large receptive field}. Our module can adopt a large receptive field while maintaining the same amount of computation. Modern vision Transformer models typically resort to sparse attention~\cite{pvt, pvtv2} or window attention~\cite{swin, cswin} to mitigate the computation burden of Softmax attention. Benefited from linear complexity, our model can enjoy the advantages of a large, even global receptive field while maintaining the same computation.

\subsection{Implementation}


Our agent attention module can serve as a plug-in module and can be easily adopted on a variety of modern vision Transformer architectures. As a showcase, we empirically apply our method to four advanced and representative Transformer models including DeiT~\cite{deit}, PVT~\cite{pvt}, Swin~\cite{swin} and CSwin~\cite{cswin}. We also apply agent attention to Stable Diffusion~\cite{stable_diffusion} to accelerate image generation.
Detailed model architectures are shown in Appendix.

\section{Experiments}
\label{sec:experiments}

\begin{table}[t]

    \caption{ImageNet-1K classification results. The default input resolution is $224^2$, except $^{\uparrow\text{384}}$ denotes results on $384^2$ resolution. Check the Appendix for full results.}
    \label{tab:main}
    \centering 
    \begin{minipage}[t]{0.47 \linewidth}
        \resizebox{\linewidth}{!}{
        \setlength{\tabcolsep}{1.5mm}{
        \renewcommand\arraystretch{1.12}
        \begin{tabular}{l|c c c}
            \toprule
            \textbf{Method} & \textbf{\#Params} & \textbf{FLOPs}  & \textbf{Top-1 Acc.}\\
            
            \midrule
            DeiT-T~\cite{deit}  
                 & 5.7M     & 1.2G      & 72.2\\
            \rowcolor{lightgray!50} \textbf{Agent-DeiT-T} 
                 & 6.0M     & 1.2G      & \textbf{74.9\,{\tiny (+2.7)}}\\
            DeiT-S
                 & 22.1M    & 4.6G      & 79.8\\
            \rowcolor{lightgray!50} \textbf{Agent-DeiT-S} 
                 & 22.7M    & 4.4G      & \textbf{80.5\,{\tiny (+0.7)}}\\
            
            \midrule
            PVT-T~\cite{pvt}  
                 & 13.2M     & 1.9G      & 75.1\\
            \rowcolor{lightgray!50} \textbf{Agent-PVT-T} 
                 & 11.6M     & 2.0G      & \textbf{78.4\,{\tiny (+3.3)}}\\
            PVT-S 
                 & 24.5M     & 3.8G      & 79.8\\
            \rowcolor{lightgray!50} \textbf{Agent-PVT-S} 
                 & 20.6M     & 4.0G      & \textbf{82.2\,{\tiny (+2.4)}}\\
            PVT-L 
                 & 61.4M     & 9.8G      & 81.7\\
            \rowcolor{lightgray!50} \textbf{Agent-PVT-L} 
                 & 48.7M     & 10.4G     & \textbf{83.7\,{\tiny (+2.0)}}\\
            
            \bottomrule
        \end{tabular}}}
    \end{minipage}
    \begin{minipage}[t]{0.514 \linewidth}
        \resizebox{\linewidth}{!}{
        \setlength{\tabcolsep}{1.5mm}{
        \renewcommand\arraystretch{1.12}
        \begin{tabular}{l|c c c}
            \toprule
            \textbf{Method} & \textbf{\#Params} & \textbf{FLOPs}  & \textbf{Top-1 Acc.}\\
            
            \midrule
                        Swin-T~\cite{swin}  
                 & 29M       & 4.5G      & 81.3\\
            \rowcolor{lightgray!50} \textbf{Agent-Swin-T} 
                 & 29M       & 4.5G      & \textbf{82.6\,{\tiny (+1.3)}}\\
            Swin-S 
                 & 50M       & 8.7G      & 83.0\\
            \rowcolor{lightgray!50} \textbf{Agent-Swin-S} 
                 & 50M       & 8.7G      & \textbf{83.7\,{\tiny (+0.7)}}\\
            Swin-B 
                 & 88M       & 15.4G     & 83.5\\
            \rowcolor{lightgray!50} \textbf{Agent-Swin-B} 
                 & 88M       & 15.4G     & \textbf{84.0\,{\tiny (+0.5)}}\\
            
            \midrule
            CSwin-B~\cite{cswin}
                 & 78M       & 15.0G     & 84.2\\
            \rowcolor{lightgray!50} \textbf{Agent-CSwin-B} 
                 & 73M       & 14.9G     & \textbf{84.7\,{\tiny (+0.5)}}\\
            CSwin-B$^{\uparrow\text{384}}$  
                & 78M       & 47.0G     & 85.4\\
            \rowcolor{lightgray!50} \textbf{Agent-CSwin-B}$^{\uparrow\text{384}}$  
                 & 73M       & 46.3G     & \textbf{85.8\,{\tiny (+0.4)}}\\
            
            \bottomrule
        \end{tabular}}}
    \end{minipage}
\end{table}

\begin{figure}[t]
    \centering
    \includegraphics[width=1.0\linewidth]{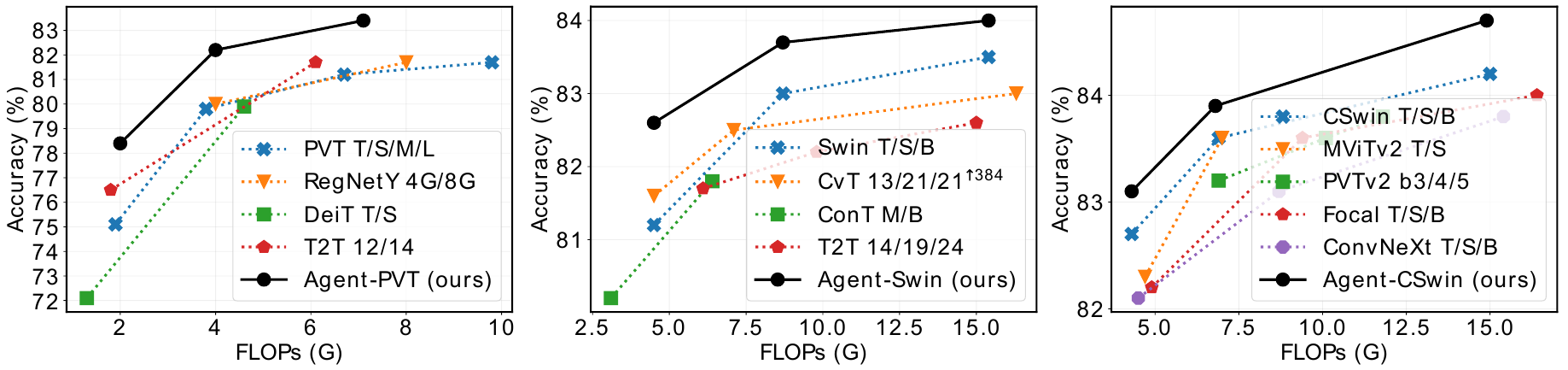}
    \caption{Comparison with SOTA models~\cite{regnety, deit, t2tvit, cvt, cont, mvitv2, pvtv2, focal, convnext} on ImageNet-1K.}
    \label{fig:main}
\end{figure}

To verify the effectiveness of our method, we conduct experiments on ImageNet-1K classification~\cite{imagenet}, ADE20K semantic segmentation~\cite{ade20k}, and COCO object detection~\cite{coco}. Additionally, we integrate agent attention into the state-of-the-art generation model, Stable Diffusion~\cite{stable_diffusion}. Furthermore, we construct high-resolution models with large receptive fields to maximize the benefits of agent attention.  In addition, sufficient ablation experiments are conducted to show the effectiveness of each design.

\begin{table}[t]
\caption{Results on COCO dataset. The FLOPs are computed over backbone, FPN and detection head with an input resolution of 1280$\times$800. See full results in Appendix.}
\label{tab:det2}
\centering 
\begin{minipage}[t]{0.495 \linewidth}
    \resizebox{\linewidth}{!}{
    \setlength{\tabcolsep}{0.5mm}{
    \renewcommand\arraystretch{1.12}
    \begin{tabular}{l|c|c|ccc|ccc}
        \toprule
        \multicolumn{9}{c}{\textbf{(a) Mask R-CNN Object Detection}} \\
        Method & FLOPs & Sch. & AP$^b$ & AP$^b_\text{50}$ & AP$^b_\text{75}$ & AP$^m$ & AP$^m_\text{50}$ & AP$^m_\text{75}$ \\
        
        \hline PVT-T 
        & 240G & 1x      & 36.7 & 59.2 & 39.3 & 35.1 & 56.7 & 37.3 \\
        \rowcolor{lightgray!50} Agent-PVT-T
        & 230G & 1x      & 41.4 & 64.1 & 45.2 & 38.7 & 61.3 & 41.6 \\
    
    
        \hline PVT-M     
        & 392G & 1x      & 42.0 & 64.4 & 45.6 & 39.0 & 61.6 & 42.1 \\
        \rowcolor{lightgray!50} Agent-PVT-M
        & 400G & 1x      & 45.9 & 67.8 & 50.4 & 42.0 & 65.0 & 45.4 \\
    
        \hline PVT-L 
        & 494G & 1x      & 42.9 & 65.0 & 46.6 & 39.5 & 61.9 & 42.5 \\
        \rowcolor{lightgray!50} Agent-PVT-L
        & 510G & 1x      & 46.9 & 69.2 & 51.4 & 42.8 & 66.2 & 46.2 \\
        
        
    
        \hline Swin-S
        & 358G & 1x      & 45.7 & 67.9 & 50.4 & 41.1 & 64.9 & 44.2 \\
        \rowcolor{lightgray!50} Agent-Swin-S
        & 364G & 1x      & 47.2 & 69.6 & 52.3 & 42.7 & 66.6 & 45.8 \\
        
        \toprule
    \end{tabular}
    }}
\end{minipage}
\begin{minipage}[t]{0.495 \linewidth}
    \resizebox{\linewidth}{!}{
    \setlength{\tabcolsep}{0.5mm}{
    \renewcommand\arraystretch{1.12}
    \begin{tabular}{l|c|c|ccc|ccc}
        \toprule        
        \multicolumn{9}{c}{\textbf{(b) Cascade Mask R-CNN Object Detection}} \\
        Method & FLOPs & Sch. & AP$^b$ & AP$^b_\text{50}$ & AP$^b_\text{75}$ & AP$^m$ & AP$^m_\text{50}$ & AP$^m_\text{75}$ \\
        
        \hline Swin-T     
        & 745G & 1x      & 48.1 & 67.1 & 52.2 & 41.7 & 64.4 & 45.0 \\
        \rowcolor{lightgray!50} Agent-Swin-T
        & 755G & 1x      & 49.2 & 68.6 & 53.2 & 42.7 & 65.6 & 45.9 \\
        
        \hline Swin-T 
        & 745G & 3x      & 50.4 & 69.2 & 54.7 & 43.7 & 66.6 & 47.3 \\
        \rowcolor{lightgray!50} Agent-Swin-T
        & 755G & 3x      & 51.4 & 70.2 & 55.9 & 44.5 & 67.6 & 48.4 \\
        
        \hline Swin-S 
        & 837G & 3x      & 51.9 & 70.7 & 56.3 & 45.0 & 68.2 & 48.8 \\
        \rowcolor{lightgray!50} Agent-Swin-S
        & 843G & 3x      & 52.6 & 71.3 & 57.1 & 45.5 & 68.9 & 49.2 \\
    
        \hline Swin-B 
        & 981G & 3x      & 51.9 & 70.5 & 56.4 & 45.0 & 68.1 & 48.9 \\
        \rowcolor{lightgray!50} Agent-Swin-B
        & 990G & 3x      & 52.6 & 71.1 & 57.1 & 45.3 & 68.6 & 49.2 \\
        
        \toprule
    \end{tabular}
    }}
\end{minipage}
\end{table}

\begin{figure}[t]
    \centering
    \includegraphics[width=1.0\linewidth]{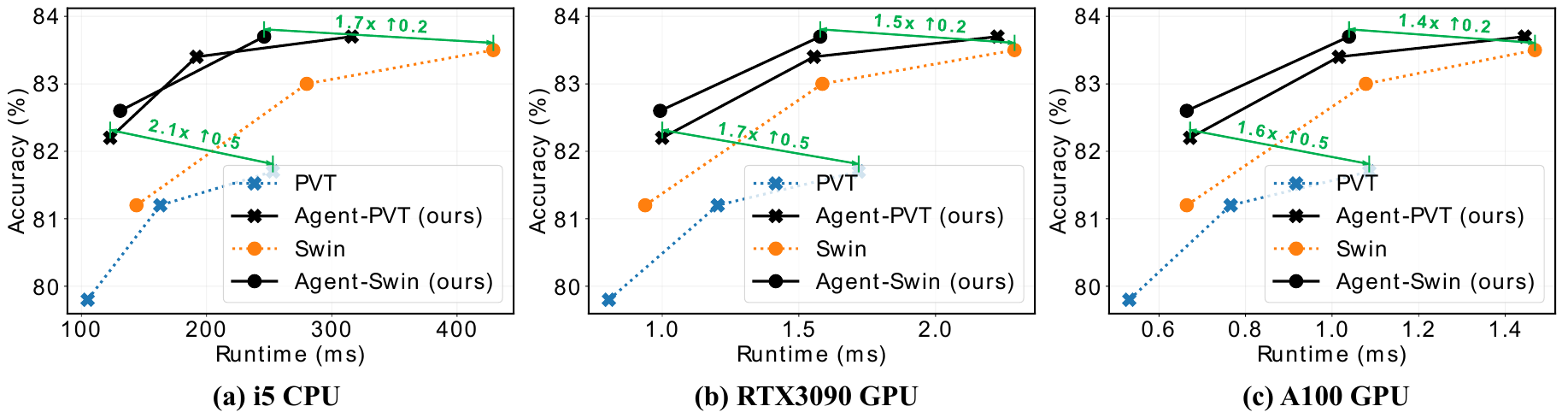}
    \caption{Accuracy-Runtime curve on ImageNet. Runtime is tested with resolution $224^2$.}
    \label{fig:speed}
\end{figure}

\subsection{ImageNet-1K Classification}

ImageNet~\cite{imagenet} comprises 1000 classes, with 1.2 million training images and 50,000 validation images. We implement our module on four representative vision Transformers and compare the top-1 accuracy on the validation split with state-of-the-art models. See Appendix for \textbf{training settings}.


\noindent \textbf{Results.}
As depicted in \cref{tab:main}, substituting Softmax attention with agent attention in various models results in significant performance improvements. For instance, Agent-PVT-S surpasses PVT-L while using just 30\% of the parameters and 40\% of the FLOPs. Additionally, we provide a comprehensive comparison with various state-of-the-art methods in \cref{fig:main}. Our models clearly achieve a better trade-off between computation cost and model performance. These results unequivocally prove that our approach has robust advantages and is adaptable to diverse architectures.

\noindent \textbf{Inference Time.}
We further conduct real speed measurements by deploying the models on various devices. As \cref{fig:speed} illustrates, our models attain inference speeds 1.7 to 2.1 times faster on the CPU while simultaneously improving performance. On RTX3090 GPU and A100 GPU, our models also achieve 1.4x to 1.7x faster inference speeds.

\subsection{Object Detection}
\label{detection}

COCO \cite{coco} object detection and instance segmentation dataset has 118K training and 5K validation images. We apply our model to RetinaNet \cite{rtn}, Mask R-CNN \cite{mrcn} and Cascade Mask R-CNN \cite{cmrcn} frameworks to evaluate the performance of our method. A series of experiments are conducted utilizing both 1x and 3x schedules with different detection heads. As depicted in \cref{tab:det2}, our model exhibits consistent enhancements across all configurations. Agent-PVT outperforms PVT models with an increase in box AP ranging from \textbf{+3.9} to \textbf{+4.7}, while Agent-Swin surpasses Swin models by up to \textbf{+1.5} box AP. These substantial improvements can be attributed to the large receptive field brought by our design, proving the effectiveness of agent attention in high-resolution scenarios.

\begin{table}[t]
\caption{Results of semantic segmentation on ADE20K. The FLOPs are computed over encoders and decoders with an input image at the resolution of 512$\times$2048.}
\label{tab:seg}
\centering 
\begin{minipage}[t]{0.495 \linewidth}
    \resizebox{\linewidth}{!}{
    \setlength{\tabcolsep}{0.5mm}{
    \renewcommand\arraystretch{1.12}
    \begin{tabular}{l|c|cc|cc}
        \toprule
        \multicolumn{6}{c}{\textbf{SemanticFPN Semantic Segmentation}} \\
        Backbone & Method & FLOPs & \#Params & mIoU & mAcc \\
        
        \hline PVT-T 
        & S-FPN & 158G & 17M & 36.57 & 46.72 \\
        \rowcolor{lightgray!50} Agent-PVT-T
        & S-FPN & 147G & 15M & \textbf{40.18} & 51.76 \\
    
        \hline PVT-S 
        & S-FPN & 225G & 28M & 41.95 & 53.02 \\
        \rowcolor{lightgray!50} Agent-PVT-S
        & S-FPN & 211G & 24M & \textbf{44.18} & 56.17 \\
    
        \hline PVT-L 
        & S-FPN & 420G & 65M & 43.49 & 54.62 \\
        \rowcolor{lightgray!50} Agent-PVT-L
        & S-FPN & 434G & 52M & \textbf{46.52} & 58.50 \\
        
        \toprule
    \end{tabular}
    }}
\end{minipage}
\begin{minipage}[t]{0.495 \linewidth}
    \resizebox{\linewidth}{!}{
    \setlength{\tabcolsep}{0.5mm}{
    \renewcommand\arraystretch{1.12}
    \begin{tabular}{l|c|cc|cc}
        \toprule
        \multicolumn{6}{c}{\textbf{UperNet Semantic Segmentation}} \\
        Backbone & Method & FLOPs & \#Params & mIoU & mAcc \\
        
        \hline Swin-T 
        & UperNet & 945G & 60M & 44.51 & 55.61 \\
        \rowcolor{lightgray!50} Agent-Swin-T
        & UperNet & 954G & 61M & \textbf{46.68} & 58.53 \\
        
        \hline Swin-S
        & UperNet & 1038G & 81M & 47.64 & 58.78 \\
        \rowcolor{lightgray!50} Agent-Swin-S
        & UperNet & 1043G & 81M & \textbf{48.08} & 59.78 \\
    
        \hline Swin-B 
        & UperNet & 1188G & 121M & 48.13 & 59.13 \\
        \rowcolor{lightgray!50} Agent-Swin-B
        & UperNet & 1196G & 121M & \textbf{48.73} & 60.01 \\
        
        \toprule
    \end{tabular}
    }}
\end{minipage}
\end{table}

\subsection{Semantic Segmentation}

ADE20K \cite{ade20k} is a well-established benchmark for semantic segmentation which encompasses 20K training images and 2K validation images. We apply our model to two exemplary segmentation models, namely SemanticFPN \cite{semfpn} and UperNet \cite{upernet}. The results are presented in \cref{tab:seg}. Remarkably, our Agent-PVT-T and Agent-Swin-T achieve \textbf{+3.61} and \textbf{+2.17} higher mIoU than their counterparts. The results show that our model is compatible with various segmentation backbones and consistently achieves improvements.

\subsection{Agent Attention for Stable Diffusion}


The advent of diffusion models makes it possible to generate high-resolution and high-quality images. However, current diffusion models mainly use the original Softmax attention with a global receptive field, resulting in huge computation cost and slow generation speed. In the light of this, we apply our agent attention to Stable Diffusion~\cite{stable_diffusion}, hoping to improve the generation speed of the model. Surprisingly, after simple adjustments, the Stable Diffusion model using agent attention, dubbed \textbf{AgentSD}, shows a significant improvement in generation speed and produces even better image quality \emph{without any extra training}.

\noindent \textbf{Applying agent attention to Stable Diffusion.}
We practically apply agent attention to ToMeSD model~\cite{tomesd}. ToMeSD reduces the number of tokens before attention calculation in Stable Diffusion, enhancing generation speed. Nonetheless, the post-merge token count remains considerable, resulting in continued complexity and latency. Hence, we replace the Softmax attention employed in ToMeSD model with our agent attention to further enhance speed. 
We experimentally find that when producing agent tokens through token merging~\cite{tome}, our agent attention can be directly applied to Stable Diffusion and ToMeSD model without any extra training. However, we are unable to apply the agent bias and DWC in this way. As a remedy, we make two simple adjustments to the agent attention, which are described in detail in Appendix. In addition, we get a significant boost by applying agent attention during early diffusion generation steps and keeping the later steps unchanged.

\begin{figure}[t]
    \centering
    \begin{subfigure}{0.44\linewidth}
        \centering
        \includegraphics[width=0.9\linewidth]{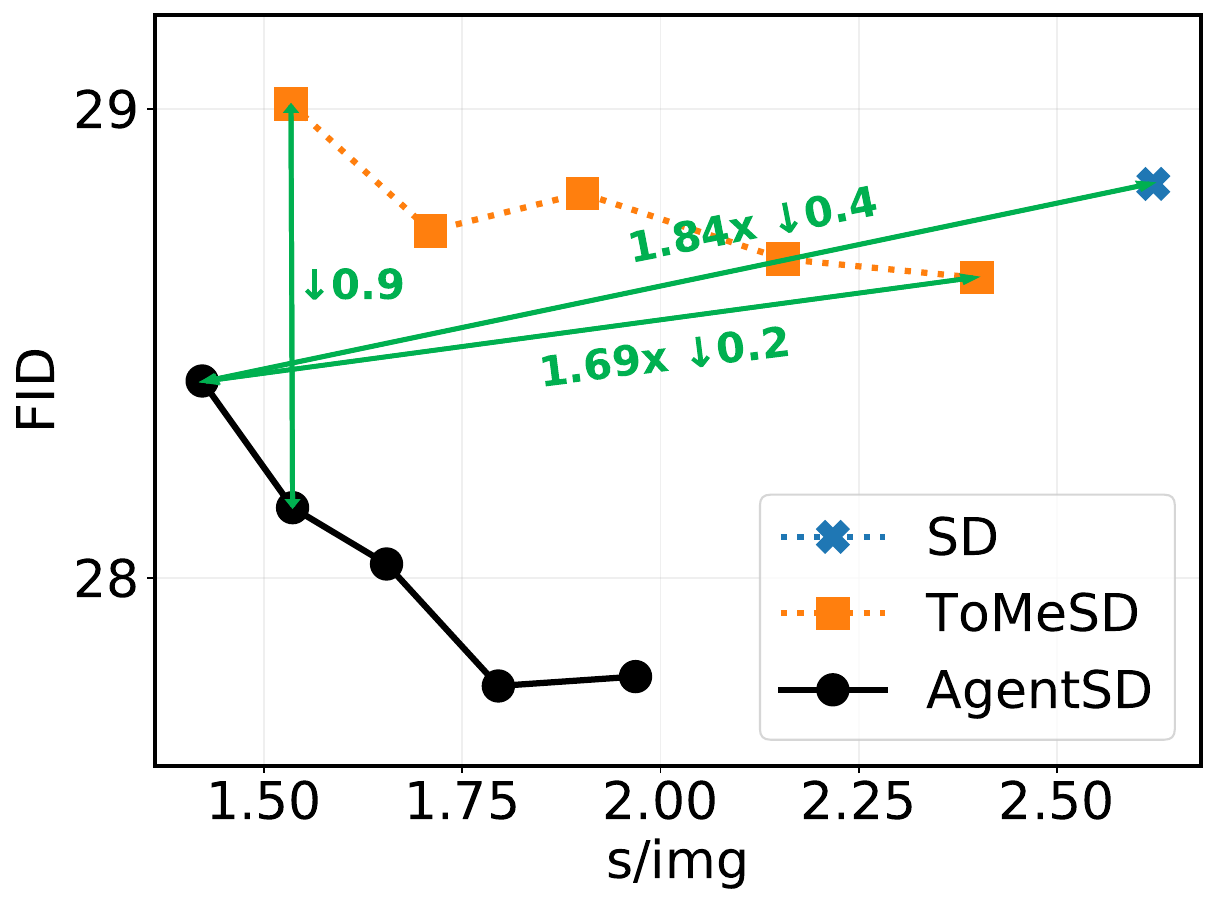}
        \caption{Quantitative results}
        \label{fig:sd_fid}
    \end{subfigure}
    \hfill
    \begin{subfigure}{0.54\linewidth}
        \includegraphics[width=0.85\linewidth]{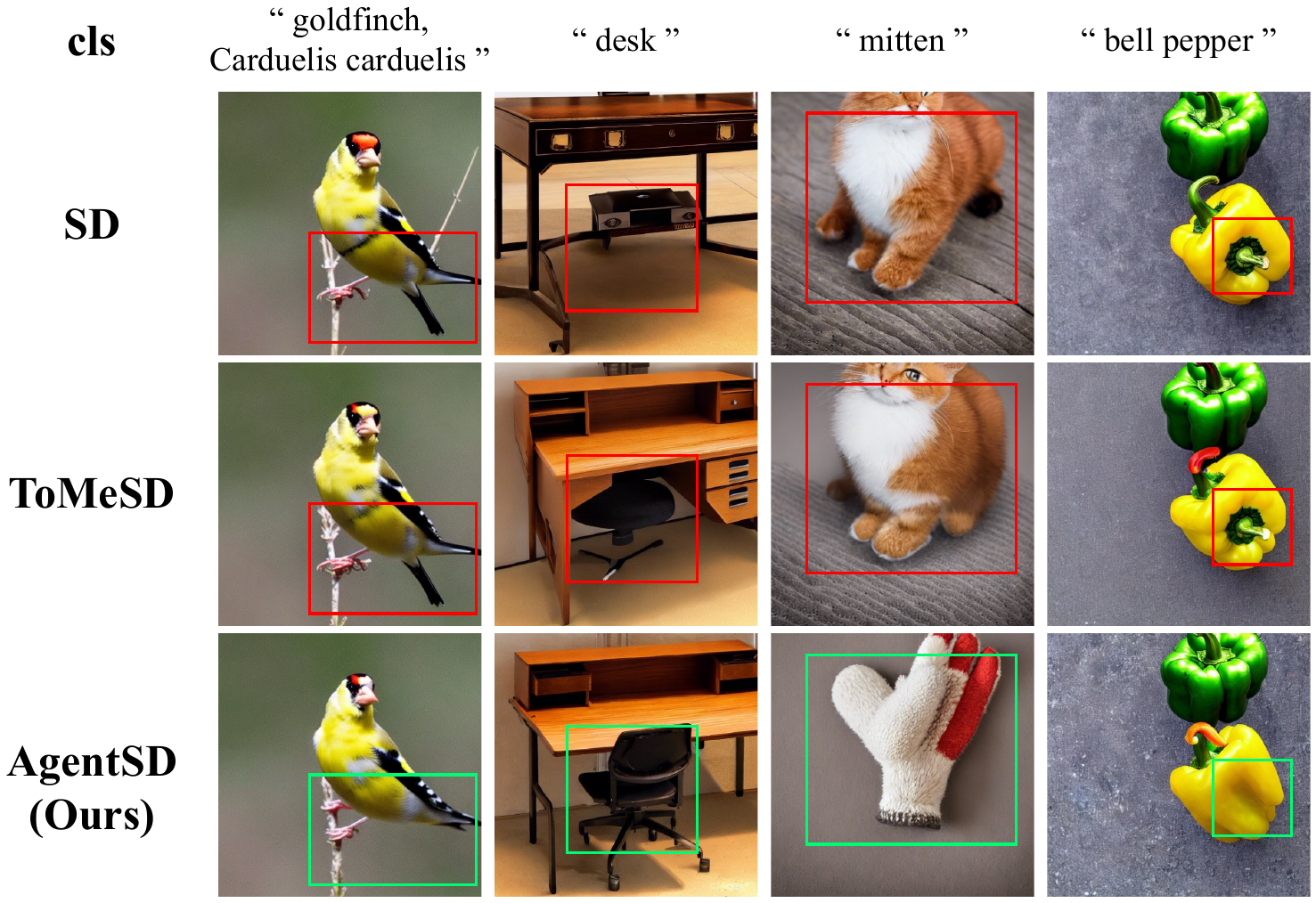}
        \caption{Samples}
        \label{fig:sd_visualize}
    \end{subfigure}
    \caption{(a) Quantitative Results of Stable Diffusion (SD), ToMeSD and our AgentSD. For ToMeSD, we take the merging ratios $\{0.1, 0.2, 0.3, 0.4, 0.5\}$ to construct five different models. Furthermore, we apply agent attention to each ToMeSD model to obtain the corresponding AgentSD model. (b) Samples generated by SD, ToMeSD ($r=40\%$) and AgentSD ($r=40\%$). The prompt is ``A high quality photograph of a \{cls\}.''.}
    \label{fig:sd_result}
\end{figure}

\noindent \textbf{Quantitative Results.}
We follow \cite{tomesd} and quantitatively compare AgentSD with Stable Diffusion and ToMeSD. 
As displayed in \cref{fig:sd_fid}, ToMeSD accelerates Stable Diffusion while maintaining similar image quality. AgentSD not only further accelerates ToMeSD but also significantly enhances image generation quality. Specifically, while maintaining superior image generation quality, AgentSD achieves 1.84x and 1.69x faster generation speeds compared to Stable Diffusion and ToMeSD, respectively. At an equivalent generation speed, AgentSD produces samples with a 0.9 lower FID score compared to ToMeSD. See the experimental details and full comparison table in Appendix.


\noindent \textbf{Visualization.}
We present some visualizations in \cref{fig:sd_visualize}. AgentSD noticeably reduces ambiguity and generation errors in comparison to Stable Diffusion and ToMeSD. For instance, in the first column, Stable Diffusion and ToMeSD produce birds with one leg and two tails, while AgentSD's sample does not exhibit this issue. In the third column, when provided with the prompt ``A high quality photo of a mitten.'', Stable Diffusion and ToMeSD erroneously generate a cat, whereas AgentSD produces the correct image.

\noindent \textbf{AgentSD for finetuning.}
We apply agent attention to SD-based Dreambooth \cite{dreambooth} to verify its performance under finetuning. When finetuned, agent attention can be integrated into all diffusion steps, reaching 2.2x acceleration in generation speed compared to the original Dreambooth. Refer to Appendix for details.

\subsection{Large Receptive Field and High Resolution} \label{sec:large_high_reso}

\begin{table}[t]
    \caption{Ablation on window size based on Agent-Swin-T.}
    \label{tab:ablation_window_size}
    \centering
    \resizebox{0.6\linewidth}{!}{
    \setlength{\tabcolsep}{1.5mm}{
    \renewcommand\arraystretch{1.2}
    \begin{tabular}{c|c c c|c c}
        \bottomrule
        \                               & Window        & FLOPs & \#Param   & Acc.      & Diff. \\
        \hline
        \multirow{4}{*}{Agent-Swin-T}   & $7^2$         & 4.5G  & 29M       & 82.0      & -0.6 \\
                                        & ${14}^2$      & 4.5G  & 29M       & 82.2      & -0.4 \\
                                        & ${28}^2$      & 4.5G  & 29M       & 82.4      & -0.2 \\
        \rowcolor{lightgray!50}
        \cellcolor{white}               & ${56}^2$      & 4.5G  & 29M       & \textbf{82.6} &\textbf{Ours} \\
        \hline
        Swin-T                          & $7^2$         & 4.5G  & 29M       & 81.3      & -1.3 \\
        \toprule
    \end{tabular}}}
\end{table}

\begin{figure}[t]
    \centering
    \includegraphics[width=1.0\linewidth]{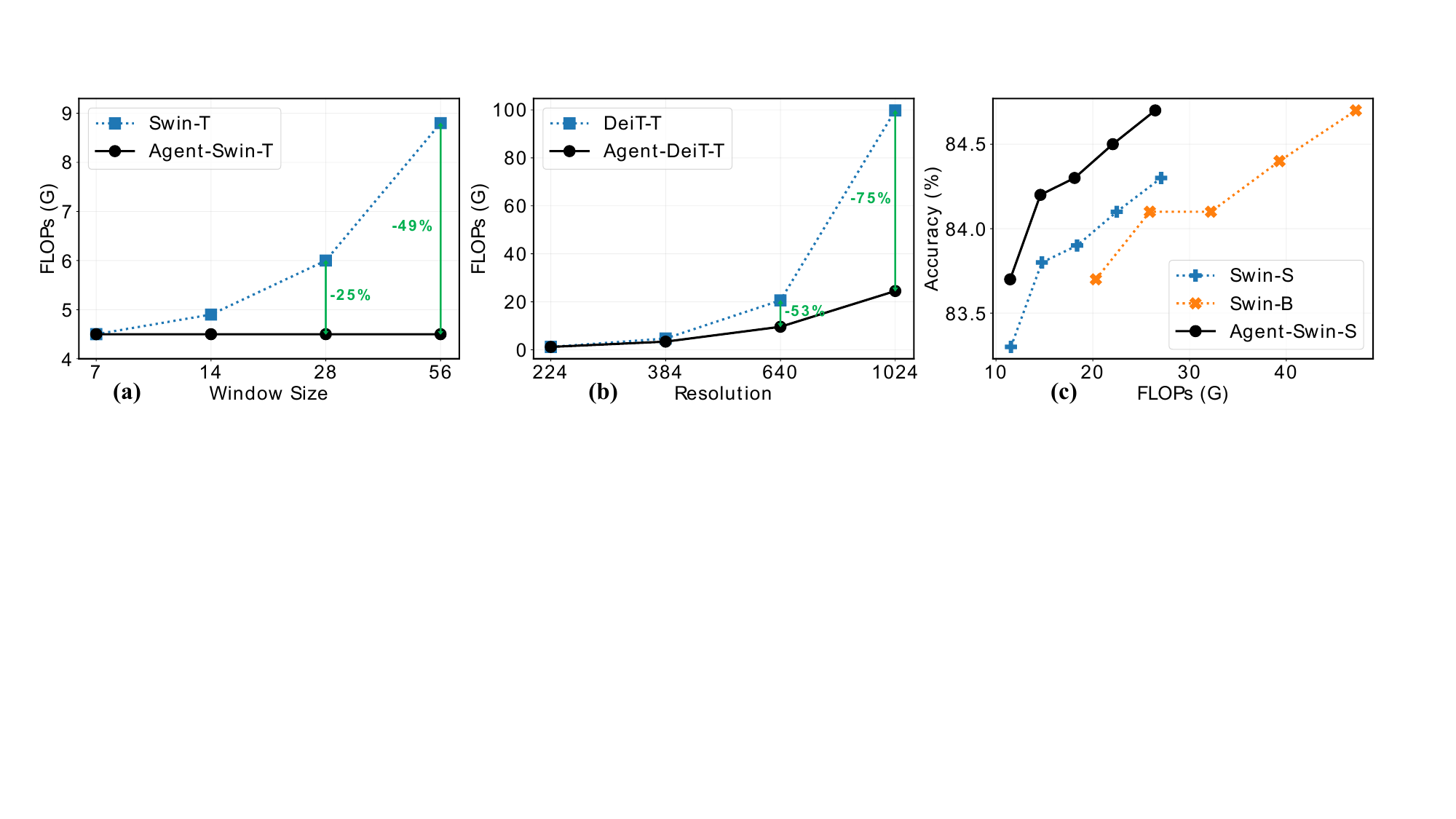}
    \caption{(a) Comparison of FLOPs between Swin and our Agent-Swin as window size increases. (b) FLOPs comparison between DeiT and our Agent-DeiT in high-resolution scenarios. (c) Increasing resolution to $\{256^2, 288^2, 320^2, 352^2, 384^2\}$. All these models are finetuned for 30 epochs from the corresponding $224^2$ resolution models.}
    \label{fig:high_reso}
\end{figure}

\begin{table}[t]
\caption{Scaling up by increasing resolution. All these models are trained from scratch.}
\label{tab:high_reso}
\centering 
\begin{minipage}[t]{0.495 \linewidth}
    \resizebox{\linewidth}{!}{
    \setlength{\tabcolsep}{0.5mm}{
    \renewcommand\arraystretch{1.1}
    \begin{tabular}{l|c c c|l}
            \toprule
            Method  & Reso  & \#Params  & Flops     & Top-1\\
            
            \midrule
            DeiT-B~\cite{deit}  
            & ${224}^2$     & 86.6M    & 17.6G     & 81.8\\
            DeiT-S  
            & ${416}^2$     & 22.2M    & 18.8G     & 82.9\,{\tiny (+1.1)}\\
            \rowcolor{lightgray!50} Agent-DeiT-B
            & ${224}^2$     & 87.2M    & 17.6G     & \textbf{82.0\,{\tiny (+0.2)}}\\
            \rowcolor{lightgray!50} Agent-DeiT-S
            & ${448}^2$     & 23.1M    & 17.7G     & \textbf{83.1\,{\tiny (+1.3)}}\\
            
            \bottomrule
    \end{tabular}
    }}
\end{minipage}
\begin{minipage}[t]{0.495 \linewidth}
    \resizebox{\linewidth}{!}{
    \setlength{\tabcolsep}{0.5mm}{
    \renewcommand\arraystretch{1.1}
    \begin{tabular}{l|c c c|l}
            \toprule
            Method  & Reso  & \#Params  & Flops     & Top-1\\
            
            \midrule
            PVT-L~\cite{pvt}  
            & ${224}^2$     & 61.4M     & 9.8G      & 81.7\\
            PVT-M
            & ${256}^2$     & 44.3M     & 8.8G      & 82.2\,{\tiny (+0.5)}\\
            \rowcolor{lightgray!50} Agent-PVT-L
            & ${224}^2$     & 48.7M     & 10.4G     & \textbf{83.7\,{\tiny (+2.0)}}\\
            \rowcolor{lightgray!50} Agent-PVT-M
            & ${256}^2$     & 36.1M     & 9.2G      & \textbf{83.8\,{\tiny (+2.1)}}\\
            
            \bottomrule
    \end{tabular}
    }}
\end{minipage}
            
            
\end{table}

\noindent \textbf{Large Receptive Field.}
Modern vision Transformers often confine self-attention calculation to local windows to reduce computation complexity, such as Swin~\cite{swin}. In \cref{tab:ablation_window_size}, we gradually enlarge the window size of Agent-Swin-T, ranging from $7^2$ to $56^2$. Clearly, as the receptive field expands, the model's performance consistently improves. This indicates that while the window attention pattern is effective, it inevitably compromises the long-range modeling capability of self-attention and remains inferior to global attention. As shown in \cref{fig:high_reso}a, unlike the quadratic complexity of Softmax attention, the linear complexity of agent attention enables us to benefit from a global receptive field while preserving identical computation complexity. 

\noindent \textbf{High Resolution.}
Limited by the quadratic complexity of Softmax attention, current vision Transformer models usually scale up by increasing model depth and width. Building on insights from \cite{efficientnet}, we discover that enhancing resolution might be a more effective approach for scaling vision Transformers, particularly those employing agent attention with global receptive fields. As shown in \cref{tab:high_reso}, Agent-DeiT-B achieves a 0.2 accuracy gain compared to DeiT-B, whereas Agent-DeiT-S at $448^2$ resolution attains an accuracy of 83.1 with only a quarter of the parameters. We observed analogous trends when scaling the resolution of Agent-PVT-M and Agent-Swin-S (see Appendix). 
\cref{fig:high_reso}b shows the FLOPs comparison between Agent-DeiT and DeiT, with Agent-DeiT saving 75\% of FLOPs for $1024^2$ resolution images.
In \cref{fig:high_reso}c, we progressively increase the resolution of Agent-Swin-S, Swin-S, and Swin-B. It is evident that in high-resolution scenarios, our model consistently delivers notably superior outcomes. 


\subsection{Ablation Study}

In this section, we ablate the key components in our agent attention module to verify the effectiveness of these designs. We report the results on ImageNet-1K classification based on Agent-Swin-T.

\begin{table}[t]
    \caption{Ablation on each module of agent attention.}
    \label{tab:ablation_agent_bias_dwc}
    \centering
    \scriptsize
    \setlength{\tabcolsep}{1.5mm}{
    \renewcommand\arraystretch{1.2}
    \begin{tabular}{l|c c|c c}
        \bottomrule
        \                           & FLOPs     & \#Param   & Acc.          & Diff. \\
        \hline
        Vanilla Linear Attention    & 4.5G      & 29M      & 77.8          & -4.8 \\
        Agent Attention             & 4.5G      & 29M      & 79.0          & -3.6 \\
        ~~~ $+$ \ Agent Bias        & 4.5G      & 29M      & 81.1          & -1.5 \\
        \rowcolor{lightgray!50}
        ~~~ $+$ \ DWC               & 4.5G      & 29M      & \textbf{82.6} & \textbf{Ours} \\
        \hline
        Swin-T w/o PE               & 4.5G      & 29M      & 80.1          & -2.5 \\
        ~~~ $+$ \ RPE               & 4.5G      & 29M      & 81.3          & -1.3 \\
        ~~~ $+$ \ DWC               & 4.5G      & 29M      & 81.6          & -1.0 \\
        \toprule
    \end{tabular}}
\end{table}

\noindent \textbf{Ablation on key designs.}
We substitute Softmax attention in Swin-T with vanilla linear attention, followed by a gradual introduction of agent attention, agent bias, and DWC to create Agent-Swin-T. The results are depicted in \cref{tab:ablation_agent_bias_dwc}. Three key findings emerge: (1) Agent attention boosts accuracy by 1.2, proving its effectiveness. (2) Agent bias serves as an effective position embedding for agent attention, similar to RPE in Swin. (3) DWC is a crucial complement to unlock the capabilities of agent attention. When applying DWC to Swin-T, a modest gain of 0.3 is observed. In contrast, with DWC preserving feature diversity, agent attention delivers a much better result (+1.5).

\noindent \textbf{Ablation on number of agent tokens.}
The model's computation complexity can be modulated by varying the number of agent tokens. As shown in \cref{tab:ablation_agent_num}, shallower layers of the model have simple semantics, and judiciously decreasing the number of agent tokens in these layers does not adversely affect performance.
In contrast, deeper layers have rich semantics, and reducing agent tokens in these layers leads to performance degradation. Hence, our design principle is using fewer agent tokens in the model's shallow layers to reduce computation complexity and more agent tokens in the deep layers to better represent rich semantics. This aligns with the stripe width design principle in CSwin~\cite{cswin}.

\begin{table}[t]
    \caption{Ablation on the number of agent tokens.}
    \label{tab:ablation_agent_num}
    \centering
    \scriptsize
    \setlength{\tabcolsep}{1.1mm}{
    \renewcommand\arraystretch{1.15}
    \begin{tabular}{c c c c|c c|c c}
        \bottomrule
        \multicolumn{4}{c|}{\#Num of Agent Tokens} & \multirow{2}{*}{FLOPS} & \multirow{2}{*}{\#Param} & \multirow{2}{*}{Acc.} & \multirow{2}{*}{Diff.} \\
        Stage1      & Stage2        & Stage3        & Stage4        &       &       &       &    \\
        \hline
        49          & 49            & 49            & 49            & 4.7G  & 29M   & 82.6  & -0.0  \\
        \rowcolor{lightgray!50}
        9           & 16            & 49            & 49            & 4.5G  & 29M   & \textbf{82.6} & \textbf{Ours}  \\
        9           & 16            & 25            & 49            & 4.5G  & 29M   & 82.2  & -0.4  \\
        4           & 9             & 49            & 49            & 4.5G  & 29M   & 82.4  & -0.2  \\
        \hline
        \multicolumn{4}{c|}{Swin-T}                                 & 4.5G  & 29M   & 81.3  & -1.3  \\
        \toprule
    \end{tabular}}
\end{table}

\begin{table}[t]
\caption{Comparison of different linear attention designs on DeiT-Tiny and Swin-Tiny.}
\label{tab:other_la}
\centering 
\begin{minipage}[t]{0.495 \linewidth}
    \resizebox{\linewidth}{!}{
    \setlength{\tabcolsep}{0.5mm}{
    \renewcommand\arraystretch{1.12}
    \begin{tabular}{c|c c|c}
            \bottomrule
            \multicolumn{4}{c}{\textbf{DeiT-T Setting}} \\
            Linear Attention            & FLOPs     & \#Param   & Acc. \\
            \hline
            Hydra Attn \cite{hydra_attn}
                                        & 1.1G      & 5.7M      & 68.3 \\
            Efficient Attn \cite{efficient_attn}
                                        & 1.1G      & 5.7M      & 70.2 \\
            Linear Angular Attn \cite{castling_vit}
                                        & 1.1G      & 5.7M      & 70.8 \\
            Focused Linear Attn \cite{flatten}
                                        & 1.1G      & 6.1M      & 74.1 \\
            \hline \rowcolor{lightgray!50}
            \textbf{Ours}               & 1.2G      & 6.0M      & \textbf{74.9} \\
            \toprule
    \end{tabular}
    }}
\end{minipage}
\begin{minipage}[t]{0.495 \linewidth}
    \resizebox{\linewidth}{!}{
    \setlength{\tabcolsep}{0.5mm}{
    \renewcommand\arraystretch{1.12}
    \begin{tabular}{c|c c|c}
            \bottomrule
            \multicolumn{4}{c}{\textbf{Swin-T Setting}} \\
            Linear Attention            & FLOPs     & \#Param   & Acc. \\
            \hline
            Hydra Attn \cite{hydra_attn}
                                        & 4.5G      & 29M       & 80.7 \\
            Efficient Attn \cite{efficient_attn}
                                        & 4.5G      & 29M       & 81.0 \\
            Linear Angular Attn \cite{castling_vit}
                                        & 4.5G      & 29M       & 79.4 \\
            Focused Linear Attn \cite{flatten}
                                        & 4.5G      & 29M       & 82.1 \\
            \hline
            \rowcolor{lightgray!50}
            \textbf{Ours}               & 4.5G      & 29M       & \textbf{82.6} \\
            \toprule
    \end{tabular}
    }}
\end{minipage}
\end{table}

\noindent \textbf{Comparison with Other Linear Attention.}
We conduct a comparison of our agent attention with other linear attention methods using DeiT-T and Swin-T. As depicted in \cref{tab:other_la}, substituting the Softmax attention employed by DeiT-T and Swin-T with various linear attention methods usually results in notable performance degradation. Remarkably, our models outperform all other methods as well as the Softmax baseline.

\section{Conclusion}
\label{sec:conclusion}

This paper presents a new attention paradigm dubbed \textit{Agent Attention}, which is applicable across a variety of vision Transformer models. As an elegant integration of Softmax and linear attention, agent attention enjoys both high expressive power and low computation complexity. Extensive experiments on image classification, semantic segmentation, and object detection unequivocally confirm the effectiveness of our approach, particularly in high-resolution scenarios. When integrated with Stable Diffusion, agent attention accelerates image generation and substantially enhances image quality without any extra training. Due to its linear complexity with respect to the number of tokens and its strong representation power, agent attention may pave the way for challenging tasks with super long token sequences, such as video modelling and multi-modal foundation models.

\clearpage
\section*{Acknowledgement}
This work is supported in part by the National Key R\&D Program of China under Grant 2021ZD0140407, the National Natural Science Foundation of China under Grants 42327901 and 62321005.

%
%
\bibliographystyle{splncs04}
\bibliography{main}


\section*{{\large Appendix}}


\section*{A. Comparision with Related Works}
Agent attention shares some similarities with two related works, namely GPViT~\cite{gpvit} and GRL~\cite{grl}. In this section, we provide a detailed analysis on their fundamental distinctions and the superiority of our work.

Firstly, agent attention's design is novel and superior.
Agent attention introduces a set of agent tokens $A$ to act as the ``agent'' for all queries $Q$, where $A$ is usually directly acquired from the query space, i.e.$\ A\!=\!f(Q)$. In contrast, GRL~\cite{grl} uses anchors $A$ projected from the input $X$, i.e.$\ A\!=\!\mathrm{Proj}(X)$, to compute cross-scale similarity (see its Fig.~6). GPViT~\cite{gpvit} uses learnable tokens and additional MLPs to achieve global modeling (see its Fig.~3). As a result, agent attention can be applied to existing models in a training-free manner, \textit{e.g.}, our AgentSD, which is impractical for GRL~\cite{grl} and GPViT~\cite{gpvit} since they need extra training of the projections or MLPs. 

Secondly, our perspective of generalized linear attention is unique and essential. This perspective enables us to unleash the potential of agent attention with lightweight linear attention enhancements like DWC. In contrast, without such a view, GPViT~\cite{gpvit} and GRL~\cite{grl} have to compromise with other techniques, such as heavy MLP, window attention, or channel attention, to achieve comparable results.

Thirdly, agent attention is a superior alternative to Softmax attention, acting as a versatile module for general purposes, while GPViT~\cite{gpvit} and GRL~\cite{grl} are not plug-in modules and are limited to specific tasks.

\section*{B. Composition of Agent Bias}
As mentioned in the main paper, to better utilize positional information, we present a carefully designed \textit{Agent Bias} for our agent attention, i.e.,
\begin{equation} \label{eq:agnet_bias}
    \begin{split}
        O^{\rm A}\!=\ &{\rm \sigma}(QA^T\!\!+\!B_2)\ {\rm \sigma}(AK^T\!\!+\!B_1)\ V, \\
    \end{split}
\end{equation}
where $ B_1\in\mathbb{R}^{n \times N},B_2\in\mathbb{R}^{N \times n} $ are our agent biases. For parameter efficiency, we construct each agent bias using three bias components rather than directly setting $B_1, B_2$ as learnable parameters. For instance, values in $B_1$ are derived from column bias $B_{1c}\in\mathbb{R}^{n \times 1 \times w}$, row bias $ B_{1r}\in\mathbb{R}^{n \times h \times 1}$ and block bias $ B_{1b}\in\mathbb{R}^{n \times h_0 \times w_0}$, where $h, w$ are the height and width of feature map or attention window, while $h_0, w_0$ are predefined hyperparameters much smaller than $h, w$. During the computation of attention weights, $B_{1c}, B_{1r}, B_{1b}$ are resized to $B'_{1c}, B'_{1r}, B'_{1b}\in\mathbb{R}^{n \times h \times w}$ by repeating or interpolating, and $ B_1=(B'_{1c}+B'_{1r}+B'_{1b}).{\rm reshape}(n, N) $ is used as the full agent bias.

\section*{C. Agent Attention for Stable Diffusion}
\subsection*{C.1. Adjustments}
As discussed in the main paper, when producing agent tokens through token merging, our agent attention can be directly applied to the Stable Diffusion model without any extra training. However, we are unable to apply the agent bias and DWC without training. As a remedy, we make two simple adjustments to our agent attention. Moreover, we change our agent attention module from
\begin{equation} \label{eq:agent_attn_module}
    \begin{split}
        O=&\ {\rm \sigma}(QA^T\!\!+\!B_2)\ {\rm \sigma}(AK^T\!\!+\!B_1)\ V+{\rm DWC}(V),
    \end{split}
\end{equation}
to
\begin{equation} \label{eq:agent_attn_for_sd}
    \begin{split}
        O=&\ {\rm \sigma}(QA^T)\ {\rm \sigma}(AK^T)\ V+kV,
    \end{split}
\end{equation}
where $k$ is a predefined hyperparameter. On the other hand, compared to the original softmax attention, the two softmax attention operations of agent attention may result in smoother feature distribution without training. In the light of this, we slightly increase the scale used for the second Softmax attention, i.e., agent broadcast. 

\subsection*{C.2. Experiment Details}
To quantitatively compare AgentSD with Stable Diffusion and ToMeSD, we follow \cite{tomesd} and employ Stable Diffusion v1.5 to generate 2,000 $512^2$ images of ImageNet-1k~\cite{imagenet} classes, featuring two images per class, using 50 PLMS~\cite{plms} diffusion steps with a cfg scale~\cite{diffusion} of 7.5. Subsequently, we calculate FID~\cite{gans} scores between these 2,000 samples and 50,000 ImageNet-1k validation examples, employing \cite{fid}. To assess speed, we calculate the average generation time of all 2,000 samples on a single RTX4090 GPU.

Complete quantitative results are presented in \cref{tab:agentsd_full}. Compared to SD and ToMeSD, our AgentSD not only accelerates generation and reduces memory usage, but also significantly improves image generation quality.

\begin{table}[t]
    \caption{Quantitative Results of Stable Diffusion, ToMeSD
and our AgentSD. GB/img is measured as the total memory usage change when increasing batch size by 1.}
    \label{tab:agentsd_full}
    \vskip -0.4in
    \centering
    \scriptsize
    \renewcommand\arraystretch{1.3}
    \vskip 0.4in
    \setlength{\tabcolsep}{3.0mm}{
        \begin{tabular}{l|c|c c c}
            \toprule
            \textbf{Method} 
            & \textbf{r}   & \textbf{FID} & \textbf{s/img}    & \textbf{GB/img}\\

            \midrule
            SD~\cite{stable_diffusion}  
            & 0     & 28.84     & 2.62      & 3.13\\

            \midrule
            \multirow{5}*{ToMeSD~\cite{tomesd}}
            & 0.1    & 28.64     & 2.40      & 2.55\\
            & 0.2    & 28.68     & 2.15      & 2.03\\
            & 0.3    & 28.82     & 1.90      & 2.09\\
            & 0.4    & 28.74     & 1.71      & 1.69\\
            & 0.5    & 29.01     & 1.53      & 1.47\\
            
            \midrule
            \rowcolor{lightgray!50} \cellcolor{white}
            & 0.1    & 27.79     & 1.97      & 1.77\\
            \rowcolor{lightgray!50} \cellcolor{white}
            & 0.2    & 27.77     & 1.80      & 1.60\\
            \rowcolor{lightgray!50} \cellcolor{white} \textbf{AgentSD}
            & 0.3    & 28.03     & 1.65      & 2.05\\
            \rowcolor{lightgray!50} \cellcolor{white}
            & 0.4    & 28.15     & 1.54      & 1.55\\
            \rowcolor{lightgray!50} \cellcolor{white}
            & 0.5    & 28.42     & 1.42      & 1.21\\
            \bottomrule
        \end{tabular}}
    \vskip -0.0in
\end{table}

\begin{table}[t]
    \caption{Ablation on factor $k$ of \cref{eq:agent_attn_for_sd}.}
    \label{tab:sd_ablation_shortcut}
    \vskip -0.1in
    \centering
    \scriptsize
    \setlength{\tabcolsep}{3.0mm}{
    \renewcommand\arraystretch{1.5}
    \begin{tabular}{c|c c c c}
        \bottomrule
        $k$                 & 0     & 0.025 & 0.075 & 0.15 \\
        \hline
        FID                 & 28.80 & 28.67 & \cellcolor{lightgray!50} 28.42 & 28.61 \\
        \toprule
    \end{tabular}}
    \vskip -0.1in
\end{table}

\subsection*{C.3. Ablation}
We further ablate the adjustments we made when applying agent attention to Stable Diffusion. As evident in \cref{tab:sd_ablation_shortcut} and \cref{tab:sd_ablation_scale}, both adjustments to agent attention enhance the quality of AgentSD generation. \cref{tab:sd_ablation_steps} demonstrates that applying agent attention in the early stages yields substantial performance enhancements.

\subsection*{C.4. AgentSD for finetuning}


Our agent attention module is also applicable in finetuning scenarios. To verify this, we select subject-driven task as an example and apply agent attention to SD-based Dreambooth~\cite{dreambooth}. We experimentally find that finetuning enables the integration of the agent attention module into all diffusion generation steps, reaching \textbf{2.2x} acceleration in generation speed compared to the original Dreambooth without sacrificing image quality. Additionally, time and memory cost during finetuning can be reduced as well.

\begin{table}[t]
    \caption{Ablation on scale used for the second Softmax attention.}
    \label{tab:sd_ablation_scale}
    \vskip -0.1in
    \centering
    \scriptsize
    \setlength{\tabcolsep}{3.0mm}{
    \renewcommand\arraystretch{1.5}
    \begin{tabular}{c|c c c c}
        \bottomrule
        Scale   & $d^{-0.5}$ & $d^{-0.25}$ & $d^{-0.15}$ & $d^{-0.05}$ \\
        \hline
        FID     & 28.86     & 28.64      & \cellcolor{lightgray!50} 28.42    & 28.60 \\
        \toprule
    \end{tabular}}
    \vskip -0.0in
\end{table}

\begin{table}[t]
    \caption{Ablation on how many diffusion steps to apply agent attention.}
    \label{tab:sd_ablation_steps}
    \vskip -0.1in
    \centering
    \scriptsize
    \setlength{\tabcolsep}{3.0mm}{
    \renewcommand\arraystretch{1.5}
    \begin{tabular}{c|c c c c}
        \bottomrule
        Steps   & early 20\% & early 40\% & early 60\% & early 80\% \\
        \hline
        FID     & 28.58     & \cellcolor{lightgray!50} 28.42  & 28.83    & 29.77 \\
        \hline
        s/img   & 1.50     & 1.42      & 1.39    & \cellcolor{lightgray!50} 1.34 \\
        \toprule
    \end{tabular}}
    \vskip -0.1in
\end{table}

\begin{figure*}[t]
    \centering
    \includegraphics[width=\linewidth]{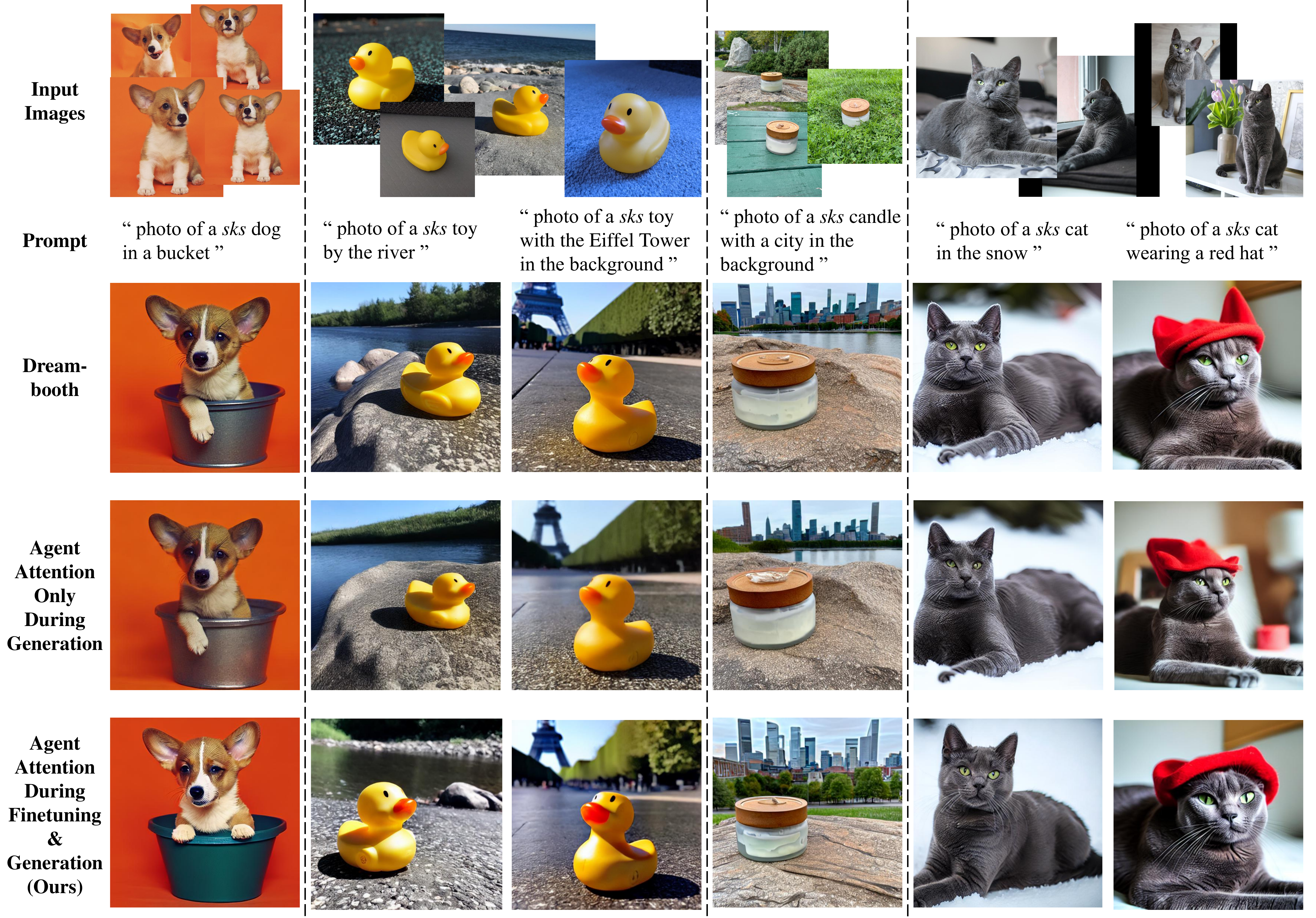}
    \vskip 0.1in
    \caption{Samples generated by Dreambooth and our Agent Dreambooth with the same seed. In the second-to-last line, we apply agent attention to all diffusion steps \textit{only during generation}, leading to a slight decline in image quality as expected. In the last row, agent attention is incorporated into all steps in \textit{both finetuning and generation}, resulting in a 2.2x speedup without compromising image quality. \textbf{Zoom in for best view.}}
    \label{fig:agent_dreambooth}
    \vskip -0.2in
\end{figure*}

\noindent \textbf{Task and baseline.} The diffusion subject-driven generation task entails maintaining the appearance of a given subject while generating novel renditions of it in different contexts, e.g., generating a photo of your pet dog dancing. Dreambooth~\cite{dreambooth} effectively addresses this task by finetuning a pretrained text-to-image diffusion model, binding a unique identifier with the given subject. Novel images of the subject can then be synthesized with the unique identifier.

\noindent \textbf{Applying agent attention to Dreambooth.} 
As previously discussed, Dreambooth\cite{dreambooth} involves an additional finetuning process. We explore two approaches to applying agent attention to Dreambooth: (1) applying it \textit{only during generation} and (2) applying it during \textit{both finetuning and generation}.
The first method is the same as the AgentSD detailed in the main paper, where we commonly apply agent attention to the early 40\% of generation steps, achieving around a 1.7x speedup (merging ratio $r=0.4$). However, applying agent attention to more diffusion steps for further acceleration leads to a decrease in image details and quality, as shown in \cref{tab:sd_ablation_steps} and the penultimate line of \cref{fig:agent_dreambooth}. Conversely, adopting the second approach, where the agent attention module is applied to all steps in both finetuning and generation, results in a 2.2x speedup in generation without sacrificing performance. Additionally, both time and memory costs in finetuning are reduced by around 15\%, enabling model finetuning with less than 12GB of GPU memory in approximately 7 minutes on a single RTX4090 GPU. The last row in \cref{fig:agent_dreambooth} shows the results of this setting. 



\noindent \textbf{Dataset and experiment details.} We adopt the dataset provided by Dreambooth~\cite{dreambooth}, which comprises 30 subjects of 15 different classes. It features live subjects and objects captured in various conditions, environments, and angles. We employ pretrained Stable Diffusion v1.5 and apply agent attention to all diffusion generation steps. The merging ratio $r$ is set to 0.4, $k$ is set to 0.075 and the scale for the second softmax attention is set to $d^{-0.15}$. We finetune all models for 800 iterations with a learning rate of 1e-6, utilizing 8-bit AdamW \cite{8bitoptim} as the optimizer. We follow \cite{dreambooth} and select \textit{sks} as the unique identifier for all settings. Novel synthesized images are sampled using the DDIM \cite{DDIM} sampler with 100 generation steps on a single RTX4090 GPU.

\noindent \textbf{Visualization and discussion.} 
Synthesized subject-driven images are shown in \cref{fig:agent_dreambooth}. We make two key observations: (1) Dreambooth with agent attention applied during finetuning and generation equals or surpasses the baseline Dreambooth in terms of fidelity and editability, and (2) employing agent attention during finetuning further enhances the fidelity and detail quality of synthesized images, enabling us to apply agent attention to all diffusion steps for more speedup. For the first observation, the first column shows that our method ensures the synthesized dog's color aligns more consistently with input images compared to the original Dreambooth and maintains comparable editability. For the second observation, comparing the last two rows of the third column reveals that applying agent attention to all diffusion steps without finetuning yields a blurry image, whereas our method produces a clearer and sharper depiction of the duck toy. Additionally, in the fifth column, our method accurately generates the cat’s eyes, whereas agent attention without finetuning fails in this aspect.

\section*{D. Dataset and Training Setup}
\subsection*{D.1. ImageNet}
\noindent \textbf{Training settings.}
To ensure a fair comparison, we train our agent attention model with the same settings as the corresponding baseline model. Specifically, we employ AdamW~\cite{adamw} optimizer to train all our models from scratch for 300 epochs, using a cosine learning rate decay and 20 epochs of linear warm-up. We set the initial learning rate to $1\times{10}^{-3}$ for a batch size of 1024 and linearly scale it \textit{w.r.t.} the batch size. Following DeiT~\cite{deit}, we use RandAugment~\cite{randaugment}, Mixup~\cite{mixup}, CutMix~\cite{cutmix}, and random erasing~\cite{random_erasing} to prevent overfitting. We also apply a weight decay of 0.05. To align with \cite{cswin}, we incorporate EMA~\cite{ema} into the training of our Agent-CSwin models. For finetuning at larger resolutions, we follow the settings in~\cite{swin, cswin} and finetune the models for 30 epochs.

\subsection*{D.2. COCO}
\noindent \textbf{Training settings.} COCO \cite{coco} object detection and instance segmentation dataset has 118K training and 5K validation images. We use a subset of 80k samples as training set and 35k for validation. Backbones are pretrained on ImageNet dataset with AdamW, following the training configurations mentioned in the original paper. Standard data augmentations including resize, random flip and normalize are applied. We set learning rate to 1e-4 and follow the 1x learning schedule: the whole network is trained for 12 epochs and the learning rate is divided by 10 at the 8th and 11th epoch respectively. For some models, we utilize 3x schedule: the network is trained for 36 epochs and the learning rate is divided by 10 at the 27th and 33rd epoch. All mAP results in the main paper are tested with input image size (3, 1333, 800).

\noindent \textbf{Numbers of agent tokens.} We use the ImageNet pretrained model as the backbone, which is trained with numbers of agent tokens $n$ set to $[9, 16, 49, 49]$ for the four stages respectively. As dense prediction tasks involve higher-resolution images than ImageNet, we appropriately increase the numbers of agent tokens to better preserve the rich information. Specifically, for all the Agent-PVT models, we assign the numbers of agent tokens for the four stages as $[144, 256, 784, 784]$, while for all Agent-Swin models, we allocate $[81, 144, 196, 49]$. We employ bilinear interpolation to adapt the agent bias to the increased numbers of agent tokens $n$. The same strategy is applied to ADE20k experiments as well.

\subsection*{D.3. ADE20K}
\noindent \textbf{Training settings.} ADE20K \cite{ade20k} is a well-established benchmark for semantic segmentation which encompasses 20K training images and 2K validation images. Backbones are pretrained on ImageNet with AdamW, following the training configurations mentioned in the original paper. For UperNet \cite{upernet}, we use AdamW to optimize, and set the initial learning rate as 6e-5 with a linear warmup of 1,500 iterations. Models are trained for 160K iterations in total. For Semantic FPN \cite{semfpn}, we optimize the models using AdamW for 40k iterations with an initial learning rate of 2e-4. We randomly resize and crop the image to 512 × 512 for training, and re-scale to have a shorter side of 512 pixels during testing.

\begin{table}[]
    \caption{Comparisons of agent attention with other vision transformer backbones on the ImageNet-1K classification task.}
    \label{tab:classification_full}
    \vskip -0.0in
    \centering
    \scriptsize
    \renewcommand\arraystretch{1.3}
    \setlength{\tabcolsep}{2.0mm}{
        \begin{tabular}{l|c c c|l}
            \toprule
            \textbf{Method} 
            & \textbf{Reso}   & \textbf{\#Params} & \textbf{Flops}    & \textbf{Top-1}\\
            
            \midrule
            DeiT-T~\cite{deit}  
            & ${224}^2$     & 5.7M     & 1.2G      & 72.2\\
            \rowcolor{lightgray!50} \textbf{Agent-DeiT-T} 
            & ${224}^2$     & 6.0M     & 1.2G      & \textbf{74.9\,{\scriptsize (+2.7)}}\\
            DeiT-S
            & ${224}^2$     & 22.1M    & 4.6G      & 79.8\\
            \rowcolor{lightgray!50} \textbf{Agent-DeiT-S} 
            & ${224}^2$     & 22.7M    & 4.4G      & \textbf{80.5\,{\scriptsize (+0.7)}}\\
            DeiT-B 
            & ${224}^2$     & 86.6M    & 17.6G     & 81.8\\
            \rowcolor{lightgray!50} \textbf{Agent-DeiT-B}
            & ${224}^2$     & 87.2M    & 17.6G     & \textbf{82.0\,{\scriptsize (+0.2)}}\\
            \rowcolor{lightgray!50} \textbf{Agent-DeiT-S}
            & ${448}^2$     & 23.1M    & 17.7G     & \textbf{83.1\,{\scriptsize (+1.3)}}\\
            
            \midrule
            PVT-T~\cite{pvt}  
            & ${224}^2$     & 13.2M     & 1.9G      & 75.1\\
            \rowcolor{lightgray!50} \textbf{Agent-PVT-T} 
            & ${224}^2$     & 11.6M     & 2.0G      & \textbf{78.4\,{\scriptsize (+3.3)}}\\
            PVT-S 
            & ${224}^2$     & 24.5M     & 3.8G      & 79.8\\
            \rowcolor{lightgray!50} \textbf{Agent-PVT-S} 
            & ${224}^2$     & 20.6M     & 4.0G      & \textbf{82.2\,{\scriptsize (+2.4)}}\\
            PVT-M
            & ${224}^2$     & 44.2M     & 6.7G      & 81.2\\
            \rowcolor{lightgray!50} \textbf{Agent-PVT-M} 
            & ${224}^2$     & 35.9M     & 7.0G      & \textbf{83.4\,{\scriptsize (+2.2)}}\\
            PVT-L 
            & ${224}^2$     & 61.4M     & 9.8G      & 81.7\\
            \rowcolor{lightgray!50} \textbf{Agent-PVT-L} 
            & ${224}^2$     & 48.7M     & 10.4G     & \textbf{83.7\,{\scriptsize (+2.0)}}\\
            \rowcolor{lightgray!50} \textbf{Agent-PVT-M}
            & ${256}^2$     & 36.1M     & 9.2G      & \textbf{83.8\,{\scriptsize (+2.1)}}\\
            
            \midrule
            Swin-T~\cite{swin}  
            & ${224}^2$     & 29M       & 4.5G      & 81.3\\
            \rowcolor{lightgray!50} \textbf{Agent-Swin-T} 
            & ${224}^2$     & 29M       & 4.5G      & \textbf{82.6\,{\scriptsize (+1.3)}}\\
            Swin-S 
            & ${224}^2$     & 50M       & 8.7G      & 83.0\\
            \rowcolor{lightgray!50} \textbf{Agent-Swin-S} 
            & ${224}^2$     & 50M       & 8.7G      & \textbf{83.7\,{\scriptsize (+0.7)}}\\
            Swin-B 
            & ${224}^2$     & 88M       & 15.4G     & 83.5\\
            \rowcolor{lightgray!50} \textbf{Agent-Swin-B} 
            & ${224}^2$     & 88M       & 15.4G     & \textbf{84.0\,{\scriptsize (+0.5)}}\\
            \rowcolor{lightgray!50} \textbf{Agent-Swin-S}
            & ${288}^2$     & 50M       & 14.6G     & \textbf{84.1\,{\scriptsize (+0.6)}}\\
            Swin-B 
            & ${384}^2$     & 88M       & 47.0G     & 84.5\\
            \rowcolor{lightgray!50} \textbf{Agent-Swin-B} 
            & ${384}^2$     & 88M       & 46.3G     & \textbf{84.9\,{\scriptsize (+0.4)}}\\
            
            \midrule
            CSwin-T~\cite{cswin} 
            & ${224}^2$     & 23M       & 4.3G      & 82.7\\
            \rowcolor{lightgray!50} \textbf{Agent-CSwin-T} 
            & ${224}^2$     & 21M       & 4.3G      & \textbf{83.1\,{\scriptsize (+0.4)}}\\
            CSwin-S 
            & ${224}^2$     & 35M       & 6.9G      & 83.6\\
            \rowcolor{lightgray!50} \textbf{Agent-CSwin-S} 
            & ${224}^2$     & 33M       & 6.8G      & \textbf{83.9\,{\scriptsize (+0.3)}}\\
            CSwin-B
            & ${224}^2$     & 78M       & 15.0G     & 84.2\\
            \rowcolor{lightgray!50} \textbf{Agent-CSwin-B} 
            & ${224}^2$     & 73M       & 14.9G     & \textbf{84.7\,{\scriptsize (+0.5)}}\\
            CSwin-B 
            & ${384}^2$     & 78M       & 47.0G     & 85.4\\
            \rowcolor{lightgray!50} \textbf{Agent-CSwin-B} 
            & ${384}^2$     & 73M       & 46.3G     & \textbf{85.8\,{\scriptsize (+0.4)}}\\
            \bottomrule
        \end{tabular}}
    \vskip 0.0in
\end{table}

\begin{figure}[t]
    \centering
    \includegraphics[width=1.0\linewidth]{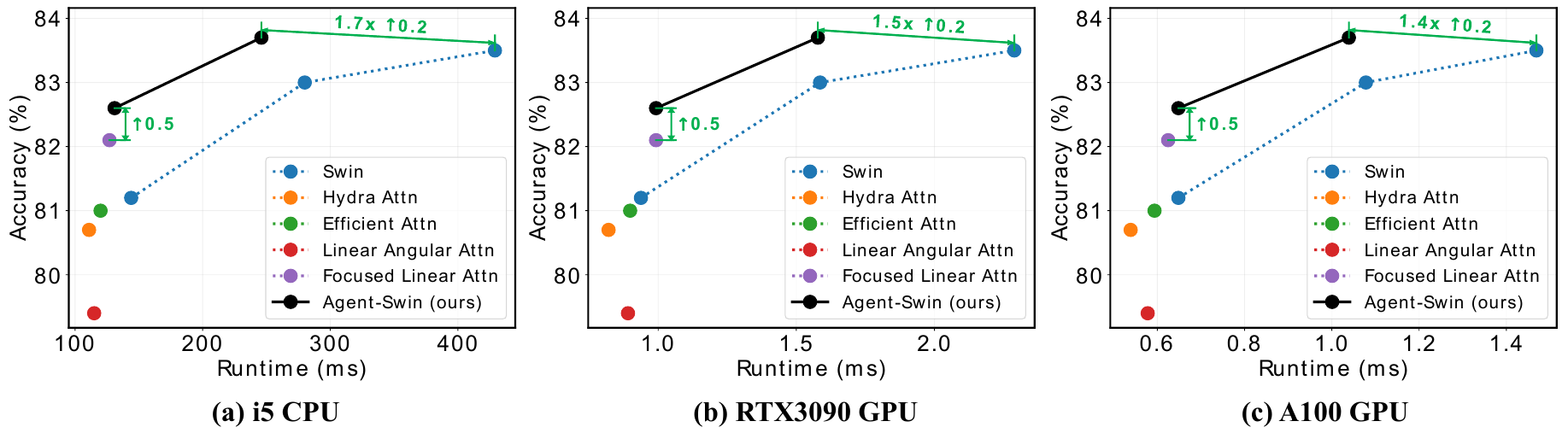}
    \vskip -0.0in
    \caption{Runtime comparison with other linear attention methods.}
    \label{fig:speed_linear}
    \vskip -0.1in
\end{figure}

\begin{table}[]
\caption{Results on COCO dataset. The FLOPs are computed over backbone, FPN and detection head with input resolution of 1280$\times$800. }
\label{tab:det4}
\vskip -0.2in
\begin{center}
\scriptsize
\setlength{\tabcolsep}{1.0mm}{
\renewcommand\arraystretch{1.3}
\begin{tabular}{l|c|c|ccc|ccc}
    \toprule
    \multicolumn{9}{c}{\textbf{(a) Mask R-CNN Object Detection on COCO}} \\
    Method & FLOPs & Sch. & AP$^b$ & AP$^b_\text{50}$ & AP$^b_\text{75}$ & AP$^m$ & AP$^m_\text{50}$ & AP$^m_\text{75}$ \\
    
    \hline PVT-T 
    & 240G & 1x      & 36.7 & 59.2 & 39.3 & 35.1 & 56.7 & 37.3 \\
    \rowcolor{lightgray!50} Agent-PVT-T
    & 230G & 1x      & 41.4 & 64.1 & 45.2 & 38.7 & 61.3 & 41.6 \\

    \hline PVT-S     
    & 305G & 1x      & 40.4 & 62.9 & 43.8 & 37.8 & 60.1 & 40.3 \\
    \rowcolor{lightgray!50} Agent-PVT-S
    & 293G & 1x      & 44.5 & 67.0 & 49.1 & 41.2 & 64.4 & 44.5 \\

    \hline PVT-M     
    & 392G & 1x      & 42.0 & 64.4 & 45.6 & 39.0 & 61.6 & 42.1 \\
    \rowcolor{lightgray!50} Agent-PVT-M
    & 400G & 1x      & 45.9 & 67.8 & 50.4 & 42.0 & 65.0 & 45.4 \\

    \hline PVT-L 
    & 494G & 1x      & 42.9 & 65.0 & 46.6 & 39.5 & 61.9 & 42.5 \\
    \rowcolor{lightgray!50} Agent-PVT-L
    & 510G & 1x      & 46.9 & 69.2 & 51.4 & 42.8 & 66.2 & 46.2 \\
    
    \hline Swin-T 
    & 267G & 1x      & 43.7 & 66.6 & 47.7 & 39.8 & 63.3 & 42.7 \\
    \rowcolor{lightgray!50} Agent-Swin-T
    & 276G & 1x      & 44.6 & 67.5 & 48.7 & 40.7 & 64.4 & 43.4 \\
    
    \hline Swin-T 
    & 267G & 3x      & 46.0 & 68.1 & 50.3 & 41.6 & 65.1 & 44.9 \\
    \rowcolor{lightgray!50} Agent-Swin-T
    & 276G & 3x      & 47.3 & 69.5 & 51.9 & 42.7 & 66.4 & 46.2 \\

    \hline Swin-S
    & 358G & 1x      & 45.7 & 67.9 & 50.4 & 41.1 & 64.9 & 44.2 \\
    \rowcolor{lightgray!50} Agent-Swin-S
    & 364G & 1x      & 47.2 & 69.6 & 52.3 & 42.7 & 66.6 & 45.8 \\

    \hline Swin-S
    & 358G & 3x      & 48.5 & 70.2 & 53.5 & 43.3 & 67.3 & 46.6 \\
    \rowcolor{lightgray!50} Agent-Swin-S
    & 364G & 3x      & 48.9 & 70.9 & 53.6 & 43.8 & 67.9 & 47.3\\
    \toprule
    
    \multicolumn{9}{c}{\textbf{(b) Cascade Mask R-CNN Object Detection on COCO}} \\
    Method & FLOPs & Sch. & AP$^b$ & AP$^b_\text{50}$ & AP$^b_\text{75}$ & AP$^m$ & AP$^m_\text{50}$ & AP$^m_\text{75}$ \\
    
    \hline Swin-T 
    & 745G & 1x      & 48.1 & 67.1 & 52.2 & 41.7 & 64.4 & 45.0 \\
    \rowcolor{lightgray!50} Agent-Swin-T
    & 755G & 1x      & 49.2 & 68.6 & 53.2 & 42.7 & 65.6 & 45.9 \\
    
    \hline Swin-T 
    & 745G & 3x      & 50.4 & 69.2 & 54.7 & 43.7 & 66.6 & 47.3 \\
    \rowcolor{lightgray!50} Agent-Swin-T
    & 755G & 3x      & 51.4 & 70.2 & 55.9 & 44.5 & 67.6 & 48.4 \\
    
    \hline Swin-S 
    & 837G & 3x      & 51.9 & 70.7 & 56.3 & 45.0 & 68.2 & 48.8 \\
    \rowcolor{lightgray!50} Agent-Swin-S
    & 843G & 3x      & 52.6 & 71.3 & 57.1 & 45.5 & 68.9 & 49.2 \\

    \hline Swin-B 
    & 981G & 3x      & 51.9 & 70.5 & 56.4 & 45.0 & 68.1 & 48.9 \\
    \rowcolor{lightgray!50} Agent-Swin-B
    & 990G & 3x      & 52.6 & 71.1 & 57.1 & 45.3 & 68.6 & 49.2 \\
    
    \toprule
\end{tabular}}
\end{center}
\vskip -0.2in
\end{table}

\begin{table}[]
\caption{Results on COCO object detection with RetinaNet \cite{rtn}. The FLOPs are computed over backbone, FPN, and detection head with an input resolution of 1280$\times$800. }
\label{tab:det3}
\vskip -0.1in
\begin{center}
\scriptsize
\setlength{\tabcolsep}{1.5mm}{
\renewcommand\arraystretch{1.3}
\begin{tabular}{l|c|ccc|ccc}
    \toprule
    \multicolumn{8}{c}{\textbf{RetinaNet Object Detection on COCO (Sch. 1x)}} \\
    Method & FLOPs & AP & AP$_\text{50}$ & AP$_\text{75}$ & AP$_{s}$ & AP$_{m}$ & AP$_{l}$ \\
    
    \hline PVT-T 
    & 221G & 36.7 & 56.9 & 38.9 & 22.6 & 38.8 & 50.0 \\
    \rowcolor{lightgray!50} Agent-PVT-T
    & 211G & 40.3 & 61.2 & 42.9 & 25.5 & 43.4 & 54.3 \\
    
    \hline PVT-S 
    & 286G &  38.7 & 59.3 & 40.8 & 21.2 & 41.6 & 54.4 \\
    \rowcolor{lightgray!50} Agent-PVT-S 
    & 274G & 44.1 & 65.3 & 47.3 & 29.2 & 47.5 & 59.8 \\
    
    \hline PVT-M 
    & 373G & 41.9 & 63.1 & 44.3 & 25.0 & 44.9 & 57.6 \\
    \rowcolor{lightgray!50} Agent-PVT-M 
    & 382G & 45.8 & 66.9 & 49.1 & 28.8 & 49.2 & 61.7 \\
    
    \hline PVT-L 
    & 475G & 42.6 & 63.7 & 45.4 & 25.8 & 46.0 & 58.4 \\
    \rowcolor{lightgray!50} Agent-PVT-L 
    & 492G & 46.8 & 68.2 & 50.7 & 30.9 & 50.8 & 62.9 \\
    
    \toprule
\end{tabular}}
\end{center}
\vskip -0.2in
\end{table}

\begin{table}[]
\caption{Results of semantic segmentation. The FLOPs are computed over encoders and decoders with an input image at the resolution of 512$\times$2048. S-FPN is short for SemanticFPN \cite{semfpn} model.}
\label{tab:seg2}
\vskip -0.1in
\begin{center}
\scriptsize
\setlength{\tabcolsep}{1.5mm}{
\renewcommand\arraystretch{1.3}
\begin{tabular}{l|c|cc|cc}
    \toprule
    \multicolumn{6}{c}{\textbf{Semantic Segmentation on ADE20K}} \\
    Backbone & Method & FLOPs & \#Params & mIoU & mAcc \\
    
    \hline PVT-T 
    & S-FPN & 158G & 17M & 36.57 & 46.72 \\
    \rowcolor{lightgray!50} Agent-PVT-T
    & S-FPN & 147G & 15M & \textbf{40.18} & 51.76 \\

    \hline PVT-S 
    & S-FPN & 225G & 28M & 41.95 & 53.02 \\
    \rowcolor{lightgray!50} Agent-PVT-S
    & S-FPN & 211G & 24M & \textbf{44.18} & 56.17 \\

    \hline PVT-M
    & S-FPN & 315G & 48M & 42.91 & 53.80 \\
    \rowcolor{lightgray!50} Agent-PVT-M
    & S-FPN & 321G & 40M & \textbf{44.30} & 56.42 \\
    
    \hline PVT-L 
    & S-FPN & 420G & 65M & 43.49 & 54.62 \\
    \rowcolor{lightgray!50} Agent-PVT-L
    & S-FPN & 434G & 52M & \textbf{46.52} & 58.50 \\
    
    \hline Swin-T 
    & UperNet & 945G & 60M & 44.51 & 55.61 \\
    \rowcolor{lightgray!50} Agent-Swin-T
    & UperNet & 954G & 61M & \textbf{46.68} & 58.53 \\
    
    \hline Swin-S
    & UperNet & 1038G & 81M & 47.64 & 58.78 \\
    \rowcolor{lightgray!50} Agent-Swin-S
    & UperNet & 1043G & 81M & \textbf{48.08} & 59.78 \\

    \hline Swin-B 
    & UperNet & 1188G & 121M & 48.13 & 59.13 \\
    \rowcolor{lightgray!50} Agent-Swin-B
    & UperNet & 1196G & 121M & \textbf{48.73} & 60.01 \\
    \toprule
\end{tabular}}
\end{center}
\vskip -0.2in
\end{table}

\begin{figure}[t]
    \centering
    \includegraphics[width=0.5\linewidth]{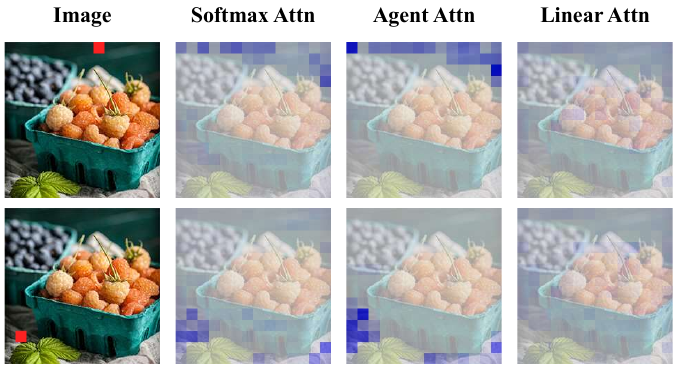}
    \vskip -0.0in
    \caption{Visualization of Softmax attention, linear attention and our agent attention. Feature corresponding to the red block is used as query.}
    \label{fig:attn_visualize}
    \vskip -0.0in
\end{figure}

\begin{figure*}[t]
    \centering
    \includegraphics[width=0.85\linewidth]{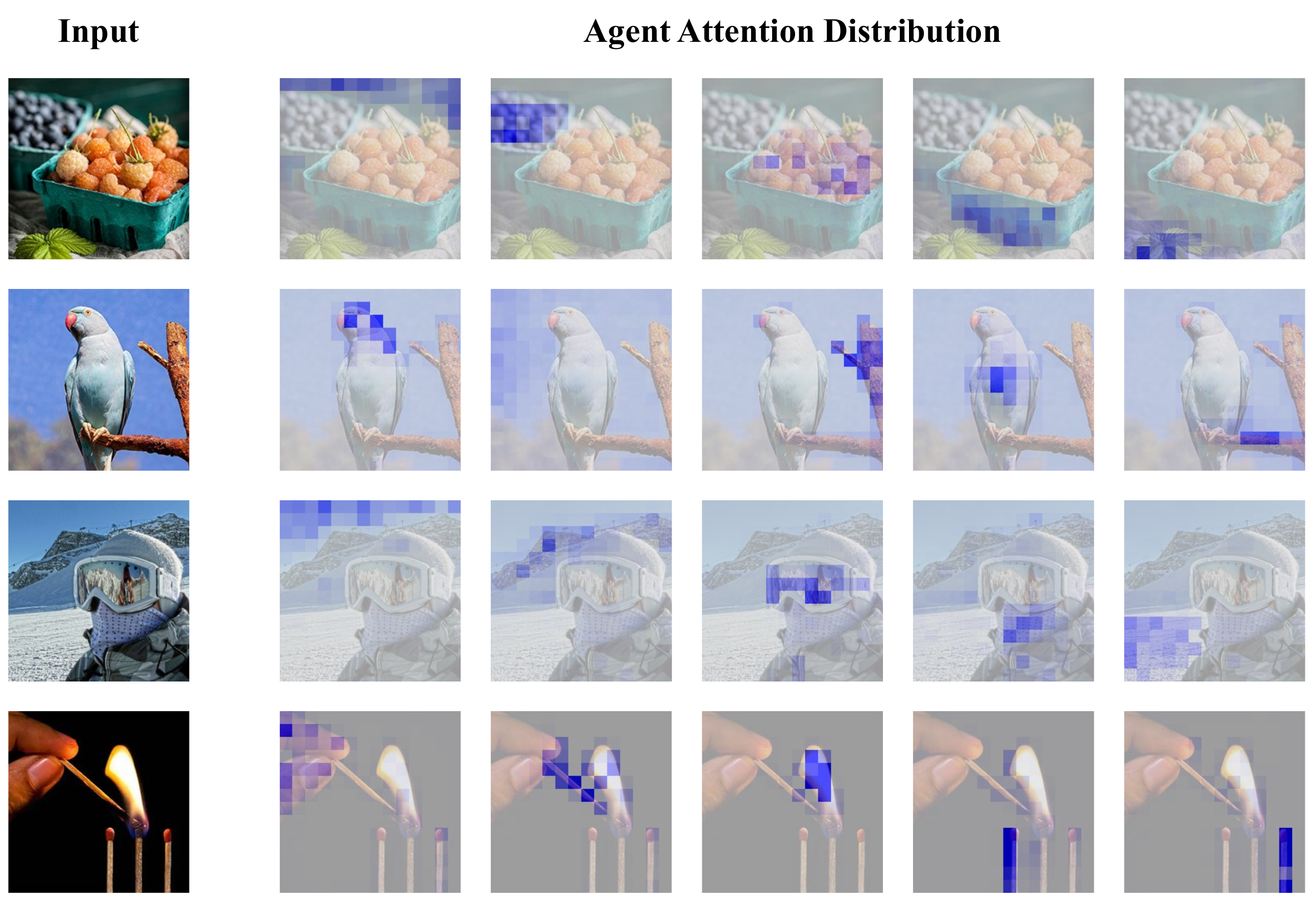}
    \vskip -0.0in
    \caption{The distribution of agent attention weights from Agent-Deit-T.}
    \label{fig:agent_visualize}
    \vskip -0.1in
\end{figure*}

\section*{E. Complete Experimental Results}

\noindent \textbf{Full classification results.}
We provide the full ImageNet-1K classification results (including high-resolution results) in \cref{tab:classification_full}. It is obvious that substituting Softmax attention with our agent attention in various models results in consistent performance improvements. We further provide additional runtime comparison with other linear attention methods in \cref{fig:speed_linear}. Agent attention achieves superior results at a comparable speed compared to other linear attention methods.

\noindent \textbf{Additional downstream experiments.}
We provide additional experiment results on object detection and semantic segmentation in Tab.\ref{tab:det3}, Tab.\ref{tab:det4} and Tab.\ref{tab:seg2}.
For object detection, results on RetinaNet \cite{rtn}, Mask R-CNN \cite{mrcn} and Cascade Mask R-CNN \cite{cmrcn} frameworks are presented, while for semantic segmentation, we show results on SemanticFPN \cite{semfpn} and UperNet \cite{upernet}. It can be observed that our models achieve consistent improvements over their baseline counterparts across various settings.

\noindent \textbf{Ablation on the type of agent tokens.}
Agent tokens can be acquired through various methods, such as setting as a set of learnable parameters or extracting from input features through pooling, depthwise convolution, etc. As shown in \cref{tab:ablation_agent_type}, dynamic agents outperform static ones due to their input-dependent nature, allowing for a more accurate representation of current queries. Pooling is a simple yet effective way to acquire agent tokens dynamically.

\begin{table}[]
    \caption{Ablation on different designs of agent tokens. }
    \label{tab:ablation_agent_type}
    \centering
    \scriptsize
    \setlength{\tabcolsep}{2.0mm}{
    \renewcommand\arraystretch{1.2}
    \begin{tabular}{l|c|c c|c c}
        \bottomrule
        Designs                 & Type      & FLOPs     & \#Param   & Acc.          & Diff. \\
        \hline
        Learnable Params        & Static    & 4.5G      & 29M       & 82.2          & -0.4 \\
        \rowcolor{lightgray!50}
        Pooling                 & Dynamic   & 4.5G      & 29M       & \textbf{82.6} & \textbf{Ours} \\
        DWC                     & Dynamic   & 4.5G      & 29M       & 82.6          & +0.0 \\
        Deformed Points~\cite{dat} & Dynamic   & 4.5G      & 29M       & 82.7          & +0.1 \\
        Token Merging~\cite{tome}  & Dynamic   & 4.6G      & 29M       & 82.6          & +0.0 \\
        \toprule
    \end{tabular}}
\end{table}

\noindent \textbf{Agent attention at different stages.}
We conduct ablation study on replacing Softmax attention with our agent attention at different stages. As depicted in \cref{tab:ablation_BBAA}, substituting the first three stages results in a performance gain of 1.3, while replacing the final stage marginally decreases overall accuracy. We attribute this outcome to the larger resolutions in the first three stages, which are more conducive to agent attention module with a global receptive field.

\begin{table}[]
    \caption{Applying agent attention module on different stages of the Swin-T structure.}
    \label{tab:ablation_BBAA}
    \centering
    \scriptsize
    \setlength{\tabcolsep}{1.1mm}{
    \renewcommand\arraystretch{1.15}
    \begin{tabular}{c c c c|c c|c c}
        \bottomrule
        \multicolumn{4}{c|}{Stages w/ Agent Attn} & \multirow{2}{*}{FLOPS} & \multirow{2}{*}{\#Param} & \multirow{2}{*}{Acc.} & \multirow{2}{*}{Diff.} \\
        Stage1      & Stage2        & Stage3        & Stage4        &       &       &       &    \\
        \hline
        $\checkmark$&               &               &               & 4.5G  & 29M   & 81.7  & -0.9  \\
        $\checkmark$&$\checkmark$   &               &               & 4.5G  & 29M   & 81.8  & -0.8  \\
        \rowcolor{lightgray!50}
        $\checkmark$&$\checkmark$   &$\checkmark$   &               & 4.5G  & 29M   & \textbf{82.6} & \textbf{Ours}  \\
        $\checkmark$&$\checkmark$   &$\checkmark$   &$\checkmark$   & 4.5G  & 29M   & 82.5  & -0.1  \\
        \hline
        \multicolumn{4}{c|}{Swin-T}                                 & 4.5G  & 29M   & 81.3  & -1.3  \\
        \toprule
    \end{tabular}}
\end{table}

\section*{F. Agent Attention Visualization}

To better understand the effectiveness of agent attention, we provide the visualization of Softmax attention, linear attention and agent attention in \cref{fig:attn_visualize}. It shows that our agent attention produces attention distributions similar to Softmax attention, while linear attention does not generate reasonable distributions. This indicates that agent attention integrates Softmax attention's expressiveness with linear complexity, resulting in its superiority.

We visualize more agent attention distributions in \cref{fig:agent_visualize}. It can be seen that various agent tokens focus on distinct regions. For instance, in the second row, agent tokens focus on head, sky, body, and branch, while the third row contains agent tokens focusing on sky, mountain, glasses, mask, and ground. This diversity ensures that different queries can focus on their areas of interest during the agent broadcast process.

\section*{G. Model Architectures}

\begin{table*}[h]
    \vskip -0.2in
    \caption{Architectures of Agent-DeiT models.}
    \label{tab:model_deit}
    \centering
    \tiny
    \setlength{\tabcolsep}{0.5mm}{
    \renewcommand\arraystretch{1.5}
    \begin{tabular}{c|c|c|c|c|c|c|c}
    \bottomrule
    \multirow{2}*{stage} & \multirow{2}*{output} & \multicolumn{2}{c|}{Agent-DeiT-T} & \multicolumn{2}{c|}{Agent-DeiT-S} & \multicolumn{2}{c}{Agent-DeiT-B}\\
    \cline{3-8}
    & & \textbf{Agent} & DeiT Block& \textbf{Agent} & DeiT Block& \textbf{Agent} & DeiT Block \\
    \hline
    \multirow{3}*[0.17in]{res1} & \multirow{3}*[0.17in]{$14\times 14$} & $\left[\!\!\! \begin{array}{c} {\rm \ win}  \ 14\!\times\! 14\ \ \\{\rm dim} \ 192 \\ {\rm head} \ 3 \\ {\rm agent} \ 49\end{array} \!\!\! \right ] \!\!\times\! 12$ & None & $\left[\!\!\! \begin{array}{c} {\rm \ win}  \ 14\!\times\! 14\ \ \\{\rm dim} \ 384 \\ {\rm head} \ 6 \\ {\rm agent} \ 49\end{array} \!\!\! \right ] \!\!\times\! 12$ & None & $\left[\!\!\! \begin{array}{c} {\rm \ win}  \ 14\!\times\! 14\ \ \\{\rm dim} \ 768 \\ {\rm head} \ 12 \\ {\rm agent} \ 81\end{array} \!\!\! \right ] \!\!\times\! 4$ & $\left[\!\!\! \begin{array}{c} {\rm \ win}  \ 14\!\times\! 14\ \ \\{\rm dim} \ 768 \\ {\rm head} \ 12\end{array} \!\!\! \right ] \!\!\times\! 8$ \\
    \toprule
    \end{tabular}}
    \vskip -0.1in
\end{table*}

\begin{table*}[h]
    \caption{Architectures of Agent-PVT models (Part1).}
    \label{tab:model_pvt-1}
    \centering
    \tiny
    \setlength{\tabcolsep}{3.5mm}{
    \renewcommand\arraystretch{1.5}
    \begin{tabular}{c|c|c|c|c|c}
    \bottomrule
    \multirow{2}*{stage} & \multirow{2}*{output} & \multicolumn{2}{c|}{Agent-PVT-T} & \multicolumn{2}{c}{Agent-PVT-S}\\
    \cline{3-6}
    & & \textbf{Agent} & PVT Block & \textbf{Agent} & PVT Block \\
    \hline
    \multirow{5}*{res1} & \multirow{5}*{$56\times 56$} & \multicolumn{4}{c}{Conv4×4, stride=4, 64, LN}\\
    \cline{3-6}
    && $\left[\!\!\! \begin{array}{c} {\rm \ win}  \ 56\!\times\! 56\ \ \\{\rm dim} \ 64 \\ {\rm head} \ 1 \\ {\rm agent} \ 9\end{array} \!\!\! \right ] \!\!\times\! 2$ & None & $\left[\!\!\! \begin{array}{c} {\rm \ win}  \ 56\!\times\! 56\ \ \\{\rm dim} \ 64 \\ {\rm head} \ 1 \\ {\rm agent} \ 9\end{array} \!\!\! \right ] \!\!\times\! 3$ & None \\
    \hline
    \multirow{5}*{res2} & \multirow{5}*{$28\times 28$} & \multicolumn{4}{c}{Conv2×2, stride=2, 128, LN}\\
    \cline{3-6}
    && $\left[\!\!\! \begin{array}{c} {\rm \ win}  \ 28\!\times\! 28\ \ \\{\rm dim} \ 128 \\ {\rm head} \ 2 \\ {\rm agent} \ 16\end{array} \!\!\! \right ] \!\!\times\! 2$ & None & $\left[\!\!\! \begin{array}{c} {\rm \ win}  \ 28\!\times\! 28\ \ \\{\rm dim} \ 128 \\ {\rm head} \ 2 \\ {\rm agent} \ 16\end{array} \!\!\! \right ] \!\!\times\! 3$ & None \\
    \hline
    \multirow{5}*{res3} & \multirow{5}*{$14\times 14$} & \multicolumn{4}{c}{Conv2×2, stride=2, 320, LN}\\
    \cline{3-6}
    & & $\left[\!\!\! \begin{array}{c} {\rm \ win}  \ 14\!\times\! 14\ \ \\{\rm dim} \ 320 \\ {\rm head} \ 5 \\ {\rm agent} \ 49\end{array} \!\!\! \right ] \!\!\times\! 2$ & None & $\left[\!\!\! \begin{array}{c} {\rm \ win}  \ 14\!\times\! 14\ \ \\{\rm dim} \ 320 \\ {\rm head} \ 5 \\ {\rm agent} \ 49\end{array} \!\!\! \right ] \!\!\times\! 6$ & None \\
    \hline
    \multirow{5}*{res4} & \multirow{5}*{$7\times 7$} & \multicolumn{4}{c}{Conv2×2, stride=2, 512, LN}\\
    \cline{3-6}
    & & $\left[\!\!\! \begin{array}{c} {\rm \ win}  \ 7\!\times\! 7\ \ \\{\rm dim} \ 512 \\ {\rm head} \ 8 \\ {\rm agent} \ 49\end{array} \!\!\! \right ] \!\!\times\! 2$ & None & $\left[\!\!\! \begin{array}{c} {\rm \ win}  \ 7\!\times\! 7\ \ \\{\rm dim} \ 512 \\ {\rm head} \ 8 \\ {\rm agent} \ 49\end{array} \!\!\! \right ] \!\!\times\! 3$ & None \\
    \toprule
    \end{tabular}}
    \vskip 0.05in
\end{table*}

\begin{table*}[h]
    \caption{Architectures of Agent-PVT models (Part2).}
    \label{tab:model_pvt-2}
    \centering
    \tiny
    \setlength{\tabcolsep}{3.5mm}{
    \renewcommand\arraystretch{1.5}
    \begin{tabular}{c|c|c|c|c|c}
    \bottomrule
    \multirow{2}*{stage} & \multirow{2}*{output} & \multicolumn{2}{c|}{Agent-PVT-M} & \multicolumn{2}{c|}{Agent-PVT-L}\\
    \cline{3-6}
    & & \textbf{Agent} & PVT Block & \textbf{Agent} & PVT Block \\
    \hline
    \multirow{5}*{res1} & \multirow{5}*{$56\times 56$} & \multicolumn{4}{c}{Conv4×4, stride=4, 64, LN}\\
    \cline{3-6}
    && $\left[\!\!\! \begin{array}{c} {\rm \ win}  \ 56\!\times\! 56 \ \ \\{\rm dim} \ 64 \\ {\rm head} \ 1 \\ {\rm agent} \ 9\end{array} \!\!\! \right ] \!\!\times\! 3$ & None & $\left[\!\!\! \begin{array}{c} {\rm \ win}  \ 56\!\times\! 56\ \ \\{\rm dim} \ 64 \\ {\rm head} \ 1 \\ {\rm agent} \ 9\end{array} \!\!\! \right ] \!\!\times\! 3$ & None \\
    \hline
    \multirow{5}*{res2} & \multirow{5}*{$28\times 28$} & \multicolumn{4}{c}{Conv2×2, stride=2, 128, LN}\\
    \cline{3-6}
    && $\left[\!\!\! \begin{array}{c} {\rm \ win}  \ 28\!\times\! 28\ \ \\{\rm dim} \ 128 \\ {\rm head} \ 2 \\ {\rm agent} \ 16\end{array} \!\!\! \right ] \!\!\times\! 3$ & None & $\left[\!\!\! \begin{array}{c} {\rm \ win}  \ 28\!\times\! 28\ \ \\{\rm dim} \ 128 \\ {\rm head} \ 2 \\ {\rm agent} \ 16\end{array} \!\!\! \right ] \!\!\times\! 8$ & None \\
    \hline
    \multirow{5}*{res3} & \multirow{5}*{$14\times 14$} & \multicolumn{4}{c}{Conv2×2, stride=2, 320, LN}\\
    \cline{3-6}
    & & $\left[\!\!\! \begin{array}{c} {\rm \ win}  \ 14\!\times\! 14\ \ \\{\rm dim} \ 320 \\ {\rm head} \ 5 \\ {\rm agent} \ 49\end{array} \!\!\! \right ] \!\!\times\! 18$ & None & $\left[\!\!\! \begin{array}{c} {\rm \ win}  \ 14\!\times\! 14\ \ \\{\rm dim} \ 320 \\ {\rm head} \ 5 \\ {\rm agent} \ 49\end{array} \!\!\! \right ] \!\!\times\! 27$ & None \\
    \hline
    \multirow{5}*{res4} & \multirow{5}*{$7\times 7$} & \multicolumn{4}{c}{Conv2×2, stride=2, 512, LN}\\
    \cline{3-6}
    & & $\left[\!\!\! \begin{array}{c} {\rm \ win}  \ 7\!\times\! 7\ \ \\{\rm dim} \ 512 \\ {\rm head} \ 8 \\ {\rm agent} \ 49\end{array} \!\!\! \right ] \!\!\times\! 3$ & None & $\left[\!\!\! \begin{array}{c} {\rm \ win}  \ 7\!\times\! 7\ \ \\{\rm dim} \ 512 \\ {\rm head} \ 8 \\ {\rm agent} \ 49\end{array} \!\!\! \right ] \!\!\times\! 3$ & None \\
    \toprule
    \end{tabular}}
    \vskip 0.05in
\end{table*}

\begin{table*}
    \caption{Architectures of Agent-Swin models.}
    \label{tab:model_swin}
    \centering
    \tiny
    \setlength{\tabcolsep}{0.5mm}{
    \renewcommand\arraystretch{1.5}
    \begin{tabular}{c|c|c|c|c|c|c|c}
    \bottomrule
    \multirow{2}*{stage} & \multirow{2}*{output} & \multicolumn{2}{c|}{Agent-Swin-T} & \multicolumn{2}{c|}{Agent-Swin-S} & \multicolumn{2}{c}{Agent-Swin-B}\\
    \cline{3-8}
    & & \textbf{Agent} & Swin Block & \textbf{Agent} & Swin Block& \textbf{Agent} & Swin Block\\
    \hline
    \multirow{5}*{res1} & \multirow{5}*{$56\times 56$} & \multicolumn{2}{c|}{concat $4\times 4$, 96, LN} & \multicolumn{2}{c|}{concat $4\times 4$, 96, LN} & \multicolumn{2}{c}{concat $4\times 4$, 128, LN}\\
    \cline{3-8}
    && $\left[\!\!\! \begin{array}{c} {\rm \ win}  \ 56\!\times\! 56\ \ \\{\rm dim} \ 96 \\ {\rm head} \ 3 \\ {\rm agent} \ 9\end{array} \!\!\! \right ] \!\!\times\! 2$ & None & $\left[\!\!\! \begin{array}{c} {\rm \ win}  \ 56\!\times\! 56\ \ \\{\rm dim} \ 96 \\ {\rm head} \ 3 \\ {\rm agent} \ 9\end{array} \!\!\! \right ] \!\!\times\! 2$ & None & $\left[\!\!\! \begin{array}{c} {\rm \ win}  \ 56\!\times\! 56\ \ \\{\rm dim} \ 128 \\ {\rm head} \ 3 \\ {\rm agent} \ 9\end{array} \!\!\! \right ] \!\!\times\! 2$ & None\\
    \hline
    \multirow{5}*{res2} & \multirow{5}*{$28\times 28$} & \multicolumn{2}{c|}{concat $2\times 2$, 192, LN} & \multicolumn{2}{c|}{concat $2\times 2$, 192, LN} & \multicolumn{2}{c}{concat $2\times 2$, 256, LN}\\
    \cline{3-8}
    && $\left[\!\!\! \begin{array}{c} {\rm \ win}  \ 28\!\times\! 28\ \ \\{\rm dim} \ 192 \\ {\rm head} \ 6 \\ {\rm agent} \ 16\end{array} \!\!\! \right ] \!\!\times\! 2$ & None & $\left[\!\!\! \begin{array}{c} {\rm \ win}  \ 28\!\times\! 28\ \ \\{\rm dim} \ 192 \\ {\rm head} \ 6 \\ {\rm agent} \ 16\end{array} \!\!\! \right ] \!\!\times\! 2$ & None & $\left[\!\!\! \begin{array}{c} {\rm \ win}  \ 28\!\times\! 28\ \ \\{\rm dim} \ 256 \\ {\rm head} \ 6 \\ {\rm agent} \ 16\end{array} \!\!\! \right ] \!\!\times\! 2$ & None\\
    \hline
    \multirow{5}*{res3} & \multirow{5}*{$14\times 14$} & \multicolumn{2}{c|}{concat $2\times 2$, 384, LN} & \multicolumn{2}{c|}{concat $2\times 2$, 384, LN} & \multicolumn{2}{c}{concat $2\times 2$, 512, LN}\\
    \cline{3-8}
    && None & $\left[\!\!\! \begin{array}{c} {\rm \ win}  \ 7\!\times\! 7\ \ \\{\rm dim} \ 384 \\ {\rm head} \ 12\end{array} \!\!\! \right ] \!\!\times\! 6$ & None & $\left[\!\!\! \begin{array}{c} {\rm \ win}  \ 7\!\times\! 7\ \ \\{\rm dim} \ 384 \\ {\rm head} \ 12\end{array} \!\!\! \right ] \!\!\times\! 18$ & $\left[\!\!\! \begin{array}{c} {\rm \ win}  \ 14\!\times\! 14\ \ \\{\rm dim} \ 512 \\ {\rm head} \ 12 \\ {\rm agent} \ 49\end{array} \!\!\! \right ] \!\!\times\! 2$ & $\left[\!\!\! \begin{array}{c} {\rm \ win}  \ 7\!\times\! 7\ \ \\{\rm dim} \ 512 \\ {\rm head} \ 12\end{array} \!\!\! \right ] \!\!\times\! 16$\\
    \hline
    \multirow{4}*{res4} & \multirow{4}*{$7\times 7$} & \multicolumn{2}{c|}{concat $2\times 2$, 768, LN} & \multicolumn{2}{c|}{concat $2\times 2$, 768, LN} & \multicolumn{2}{c}{concat $2\times 2$, 1024, LN}\\
    \cline{3-8}
    & & None& $\left[\!\!\! \begin{array}{c} {\rm \ win}  \ 7\!\times\! 7\ \ \\{\rm dim} \ 768 \\ {\rm head} \ 24\end{array} \!\!\! \right ] \!\!\times\! 2$ & None & $\left[\!\!\! \begin{array}{c} {\rm \ win}  \ 7\!\times\! 7\ \ \\{\rm dim} \ 768 \\ {\rm head} \ 24\end{array} \!\!\! \right ] \!\!\times\! 2$ & None & $\left[\!\!\! \begin{array}{c} {\rm \ win}  \ 7\!\times\! 7\ \ \\{\rm dim} \ 1024 \\ {\rm head} \ 24\end{array} \!\!\! \right ] \!\!\times\! 2$\\
    \toprule
    \end{tabular}}
\end{table*}

\begin{table*}
    \caption{Architectures of Agent-CSwin models.}
    \label{tab:model_cswin}
    \centering
    \tiny
    \setlength{\tabcolsep}{0.2mm}{
    \renewcommand\arraystretch{1.5}
    \begin{tabular}{c|c|c|c|c|c|c|c}
    \bottomrule
    \multirow{2}*{stage} & \multirow{2}*{output} & \multicolumn{2}{c|}{Agent-CSwin-T} & \multicolumn{2}{c|}{Agent-CSwin-S} & \multicolumn{2}{c}{Agent-CSwin-B}\\
    \cline{3-8}
    & & \textbf{Agent} & CSwin Block & \textbf{Agent} & CSwin Block& \textbf{Agent} & CSwin Block\\
    \hline
    \multirow{5}*{res1} & \multirow{5}*{$56\times 56$} & \multicolumn{4}{c|}{Conv7×7, stride=4, 64, LN} & \multicolumn{2}{c}{Conv7×7, stride=4, 96, LN} \\
    \cline{3-8}
    && $\left[\!\!\! \begin{array}{c} {\rm \ win}  \ 56\!\times\! 56\ \ \\{\rm dim} \ 64 \\ {\rm head} \ 2 \\ {\rm agent} \ 9\end{array} \!\!\! \right ] \!\!\times\! 2$ & None & $\left[\!\!\! \begin{array}{c} {\rm \ win}  \ 56\!\times\! 56\ \ \\{\rm dim} \ 64 \\ {\rm head} \ 2 \\ {\rm agent} \ 9\end{array} \!\!\! \right ] \!\!\times\! 3$ & None & $\left[\!\!\! \begin{array}{c} {\rm \ win}  \ 56\!\times\! 56\ \ \\{\rm dim} \ 96 \\ {\rm head} \ 4 \\ {\rm agent} \ 9\end{array} \!\!\! \right ] \!\!\times\! 3$ & None\\
    \hline
    \multirow{5}*{res2} & \multirow{5}*{$28\times 28$} & \multicolumn{4}{c|}{Conv7×7, stride=2, 128, LN} & \multicolumn{2}{c}{Conv7×7, stride=2, 192, LN} \\
    \cline{3-8}
    && $\left[\!\!\! \begin{array}{c} {\rm \ win}  \ 28\!\times\! 28\ \ \\{\rm dim} \ 128 \\ {\rm head} \ 4 \\ {\rm agent} \ 16\end{array} \!\!\! \right ] \!\!\times\! 4$ & None & $\left[\!\!\! \begin{array}{c} {\rm \ win}  \ 28\!\times\! 28\ \ \\{\rm dim} \ 128 \\ {\rm head} \ 4 \\ {\rm agent} \ 16\end{array} \!\!\! \right ] \!\!\times\! 6$ & None & $\left[\!\!\! \begin{array}{c} {\rm \ win}  \ 28\!\times\! 28\ \ \\{\rm dim} \ 192 \\ {\rm head} \ 8 \\ {\rm agent} \ 16\end{array} \!\!\! \right ] \!\!\times\! 6$ & None\\
    \hline
    \multirow{4}*{res3} & \multirow{4}*{$14\times 14$} & \multicolumn{4}{c|}{Conv7×7, stride=2, 256, LN} & \multicolumn{2}{c}{Conv7×7, stride=2, 384, LN} \\
    \cline{3-8}
    & & None & $\left[\!\!\! \begin{array}{c} {\rm \ win}  \ 7\!\times\! 14\ \ \\{\rm dim} \ 256 \\ {\rm head} \ 8\end{array} \!\!\! \right ] \!\!\times\! 18$ & None &  $\left[\!\!\! \begin{array}{c} {\rm \ win}  \ 7\!\times\! 14\ \ \\{\rm dim} \ 256 \\ {\rm head} \ 8\end{array} \!\!\! \right ] \!\!\times\! 29$ & None & $\left[\!\!\! \begin{array}{c} {\rm \ win}  \ 7\!\times\! 14\ \ \\{\rm dim} \ 384 \\ {\rm head} \ 16\end{array} \!\!\! \right ] \!\!\times\! 29$\\
    \hline
    \multirow{4}*{res4} & \multirow{4}*{$7\times 7$} & \multicolumn{4}{c|}{Conv7×7, stride=2, 512, LN} & \multicolumn{2}{c}{Conv7×7, stride=2, 768, LN} \\
    \cline{3-8}
    & & None& $\left[\!\!\! \begin{array}{c} {\rm \ win}  \ 7\!\times\! 7\ \ \\{\rm dim} \ 512 \\ {\rm head} \ 16\end{array} \!\!\! \right ] \!\!\times\! 1$ & None & $\left[\!\!\! \begin{array}{c} {\rm \ win}  \ 7\!\times\! 7\ \ \\{\rm dim} \ 512 \\ {\rm head} \ 16\end{array} \!\!\! \right ] \!\!\times\! 2$ & None & $\left[\!\!\! \begin{array}{c} {\rm \ win}  \ 7\!\times\! 7\ \ \\{\rm dim} \ 768 \\ {\rm head} \ 32\end{array} \!\!\! \right ] \!\!\times\! 2$\\
    \toprule
    \end{tabular}}
\end{table*}

We present the architectures of four Transformer models used in the main paper, including Agent-DeiT, Agent-PVT, Agent-Swin and Agent-CSwin in Tab.\ref{tab:model_deit}-\ref{tab:model_cswin}. Considering the advantage of enlarged receptive field, we mainly replace Softmax attention blocks with our agent attention module at early stages of vision Transformer models.

\end{document}